\documentclass[sigconf, nonacm]{acmart}
\usepackage{algorithm}
\usepackage{algorithmic}
\usepackage[inkscapeformat=png]{svg}
\usepackage{adjustbox}
\usepackage{multirow}
\usepackage{subcaption}
\newcommand\vldbdoi{XX.XX/XXX.XX}
\newcommand\vldbpages{XXX-XXX}
\newcommand\vldbvolume{14}
\newcommand\vldbissue{1}
\newcommand\vldbyear{2020}
\newcommand\vldbauthors{\authors}
\newcommand\vldbtitle{\shorttitle} 
\newcommand\vldbavailabilityurl{URL_TO_YOUR_ARTIFACTS}
\newcommand\vldbpagestyle{plain} 

\begin{document}
\title{Enhancing Representation Learning for Periodic Time Series with Floss: A Frequency Domain Regularization Approach}

\author{Chunwei Yang}
\affiliation{%
  \institution{Sichuan University}
  \streetaddress{Chuanda Road}
  \city{Chengdu}
  \state{China}
  \postcode{610207}
}
\email{ycwcw123@gmail.com}

\author{Xiaoxu Chen}
\orcid{0000-0003-3629-6109}
\affiliation{%
  \institution{McGill University}
  \streetaddress{817 Sherbrooke Street West}
  \city{Montreal}
  \country{Canada}
}
\email{xiaoxu.chen@mail.mcgill.ca}

\author{Lijun Sun}
\orcid{0000-0001-9488-0712}
\affiliation{%
  \institution{McGill University}
  \streetaddress{817 Sherbrooke Street West}
  \city{Montreal}
  \country{Canada}
}
\email{lijun.sun@mcgill.ca}

\author{Hongyu Yang}
\orcid{0000-0003-0030-7857}
\affiliation{%
  \institution{Sichuan University}
  \city{Chengdu}
  \country{China}
}
\email{yanghongyu@scu.edu.cn}

\author{Yuankai Wu}
\orcid{0000-0003-4435-9413}
\affiliation{%
  \institution{Sichuan University}
  \city{Chengdu}
  \country{China}
}
\email{wuyk0@scu.edu.cn}

\begin{abstract}

Time series analysis is a fundamental task in various application domains, and deep learning approaches have demonstrated remarkable performance in this area. However, many real-world time series data exhibit significant periodic or quasi-periodic dynamics that are often not adequately captured by existing deep learning-based solutions. This results in an incomplete representation of the underlying dynamic behaviors of interest. To address this gap, we propose an unsupervised method called Floss that automatically regularizes learned representations in the frequency domain. The Floss method first automatically detects major periodicities from the time series. It then employs periodic shift and spectral density similarity measures to learn meaningful representations with periodic consistency. In addition, Floss can be easily incorporated into both supervised, semi-supervised, and unsupervised learning frameworks. We conduct extensive experiments on common time series classification, forecasting, and anomaly detection tasks to demonstrate the effectiveness of Floss. We incorporate Floss into several representative deep learning solutions to justify our design choices and demonstrate that it is capable of automatically discovering periodic dynamics and improving state-of-the-art deep learning models.

\end{abstract}

\maketitle

\pagestyle{\vldbpagestyle}
\begingroup\small\noindent\raggedright\textbf{:}\\
\vldbauthors. \vldbtitle. \vldbvolume(\vldbissue): \vldbpages, \vldbyear.\\
\href{https://doi.org/\vldbdoi}{doi:\vldbdoi}
\endgroup
\begingroup
\renewcommand\thefootnote{}\footnote{\noindent
Corresponding author (Yuankai Wu, wuyk0@scu.edu.cn)
}\addtocounter{footnote}{-1}\endgroup

\ifdefempty{\vldbavailabilityurl}{}{
\vspace{.3cm}
\begingroup\small\noindent\raggedright\textbf{}\\
The source code, data, and/or other artifacts have been made available at \url{https://github.com/AgustDD/Floss}.
\endgroup
}

\section{Introduction}

We are witnessing continued developments in sensor technologies, where sensors produce multivariate time series. These advances have paved the way for the critical role of time series analysis in various scientific and engineering fields. In the realm of energy management, time series analysis enables accurate load forecasting, facilitating efficient resource allocation and optimal energy utilization~\cite{de200625, hong2016probabilistic}. Within transportation engineering, time series analysis plays a pivotal role in predicting traffic flows and optimizing transportation systems~\cite{li2018diffusion, wu2018hybrid, wu2019graph}. Moreover, in financial markets, time series analysis is of utmost importance. It allows for the modeling of asset prices, enables volatility forecasting, and assists in developing effective risk management strategies~\cite{tsay2005analysis}. The application of time series analysis in healthcare proves invaluable as well, aiding in patient monitoring, disease surveillance, and the prediction of health outcomes~\cite{lipton2016modeling, song2018attend}.

The widespread adoption of deep neural networks in time series analysis has brought about significant advancements in recent years~\cite{salinas2020deepar}. These models have demonstrated their efficacy in capturing complex temporal patterns by leveraging supervised or unsupervised training approaches. Through proper training method, neural networks acquire robust temporal representations that are well-suited for various tasks within time series analysis~\cite{yue2022ts2vec}. One crucial task where neural networks excel is forecasting, where they leverage their learned temporal representations to make accurate predictions about future values~\cite{wu2021autocts}. Additionally, neural networks have shown promising results in anomaly detection within time series data~\cite{paparrizos2022tsb}.

The quest for a universal representation of time series data has sparked significant interest in deep representation learning strategies, including contrastive learning \cite{chen2020simple, gutmann2010noise}. These strategies aim to extract powerful representations from the hidden layers of deep neural networks, capturing the intrinsic features embedded within time series data. The value of such representations extends to various downstream tasks, including time series anomaly detection, forecasting, and classification. Researchers have explored specific invariances within time series data to enhance deep representation learning frameworks. For instance, Franceschi et al.~\cite{franceschi2019unsupervised} encouraged representations that closely resemble sampled subseries, while Tonekaboni et al. ~\cite{tonekaboni2020unsupervised} enforced smoothness between adjacent time windows. Eldele et al.~\cite{ijcai2021-324} proposed a model that learns scale and permutation-invariant representations. Yue et al.~\cite{yue2022ts2vec} introduced TS2Vec, a contrastive learning framework that captures contextual invariances at multiple resolutions within time series. Despite the progress made, existing methods often borrow ideas directly from contrastive learning methods in computer vision and natural language processing domains. However, unlike images that typically possess recognizable features, time series data often exhibit underlying patterns that are not easily explainable. Applying assumptions borrowed from other domains without careful consideration may result in unsuccessful representation learning for time series data.

The temporal dynamics of real-world processes often exhibit recurring cycles and significant periodicity, which are fundamental characteristics of time series data~\cite{fuller2009introduction}. This inherent property becomes particularly evident in time series associated with human behavior, where prominent daily and weekly patterns emerge. Recognizing the importance of capturing and leveraging periodicity, the exploration of representation learning methods that effectively capture the underlying periodic invariance holds substantial promise in time series analysis. One classical approach to detecting periodicity in traditional time series analysis is the employment of frequency domain methods, which enable the identification of periodic patterns by transforming time series into the frequency domain~\cite{mcsweeney2006comparison}. The discrete Fourier transform (DFT), for instance, facilitates the conversion of time series from the time domain to the frequency domain, yielding the periodogram that encodes the strength at different frequencies. Similarly, other transformations, such as the discrete cosine transform and wavelet transform, can also identify periodicity and enhance supervised learning in time series analysis~\cite{wen2021robustperiod}. These studies provide compelling evidence that frequency domain information harbors valuable insights for analyzing periodic time series data.

In fact, frequency-domain information has been widely leveraged in deep learning architectures for modeling time series data. Zhou et al.~\cite{zhou2022fedformer} proposed the use of Transformers operating in the frequency domain, enabling the capture of global properties within time series data. Woo et al.~\cite{woo2022etsformer} introduced ETSTransformer, which utilizes Fourier bases in the frequency domain to extract dominant seasonal patterns from time series. Liu et al.~\cite{minhao2021t} devised a tree-structured network that iteratively decomposes input signals into various frequency subbands. Zhang et al.~\cite{zhang2022first} decomposed time series into seasonal and trend components, employing Fourier attention for prediction. Wu et al.~\cite{wu2023timesnet} employed Fourier transformation to disentangle original temporal variations into intraperiod and interperiod variations, capturing their dependencies using 2D convolutional operations. Notably, recent efforts have focused on regulating frequency-domain representations through unsupervised learning approaches. Zhang et al.~\cite{zhang2022self} directly applied contrastive learning to the frequency transformation of raw signals, embedding the time-based neighborhood of an example close to its frequency-based neighborhood. Similar ideas were explored in CoST~\cite{woo2021cost} and BTSF~\cite{yang2022unsupervised}. However, while these approaches leverage unsupervised learning and contrastive learning, none of them are specifically designed to capture periodic dynamics in time series data.

\begin{figure}
  \centering
  \includegraphics[width=0.75\linewidth]{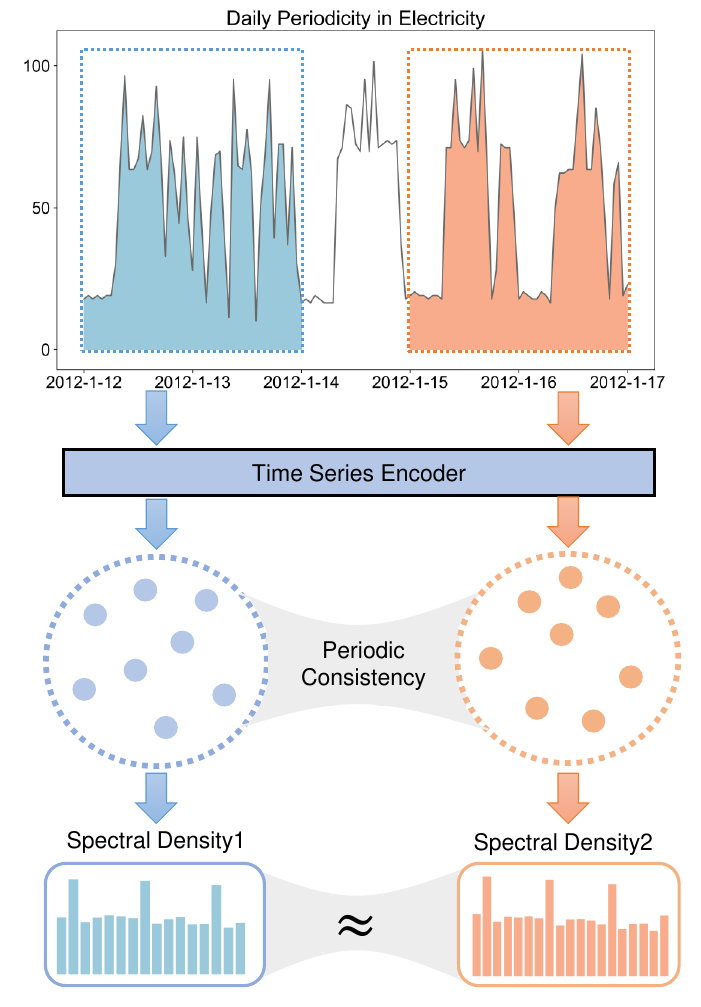}
  \caption{The framework of the paper: The time series shown in the figure exhibits strong daily periodicity. After detecting this periodicity, we aim to make the spectral densities of the representations of two time series segments, which differ by several number of periodicities, as similar as possible.}
  \label{fig:intuition}
\end{figure}

In our pursuit of capturing periodic dynamics by time series representations, we propose a novel approach that leverages the principles of contrastive learning \cite{khosla2020supervised, chen2020simple}. Contrastive learning operates on the basis of two key elements: (\romannumeral1) a contrastive loss that compares features and (\romannumeral2) a set of transformations that encode the desired invariances. Building upon this framework, we introduce a simple yet effective combination of loss function and transformation named Floss, which can be seamlessly integrated into unsupervised and semi-supervised learning methods specifically designed for periodic time series analysis. Our approach centers on the hypothesis that {\textbf{the spectral density of the learned representation remains invariant under periodic transformations}}. To realize this, our framework incorporates straightforward and efficient data augmentations that can accommodate various periodic time series with specified periodicities. Initially, we employ frequency domain transformation to automatically detect the dominant periodicity and create a periodic view of the target time series by introducing random periodic shifts in the temporal dimension. Subsequently, a time series encoder is employed to learn a periodic invariant representation. Importantly, this encoder can be seamlessly integrated into any existing deep learning framework, thereby ensuring compatibility and flexibility in its application. Finally, we design a novel task that enforces the similarity of spectral densities between the target time series and its periodic views. To mitigate the influence of high-frequency noise, we employ a hierarchical approach to measure the similarity of spectral densities between the representations. The intuition of our work is illustrated in Figure~\ref{fig:intuition}.

To the best of our knowledge, this study represents the first systematic investigation into the learning of representations for periodic or quasi-periodic time series by examining the invariance of spectral density. Specifically, our {Floss} can be seamlessly integrated into current supervised and unsupervised frameworks. The outcomes obtained from tasks such as time series classification, forecasting, and anomaly detection confirm the ability of {Floss} to capture and encode periodic invariances in time series, resulting in a notable enhancement of task performance. 

The paper is organized as follows. Section~\ref{sec:pre} introduces the necessary concepts for understanding the Floss system. In Section~\ref{sec:method}, we provide a comprehensive description of our Floss framework. Furthermore, Section~\ref{sec:exp} showcases the results of our forecasting, classification, and anomaly detection experiments using the Floss-enhanced models on extensive benchmarking datasets. In addtion, in-depth analysis and ablation studies are also provided in Section~\ref{sec:exp}. Finally, Section~\ref{sec:con} offers concluding remarks and summarizes our work.
\section{Preliminaries}
\label{sec:pre}

\noindent {\textbf{Periodic time series:}} Given a data set of periodic time series, denoted $\mathcal{X} \in \mathbb{R}^{N \times T \times F}$, where $N$ represents the number of time series and $T$ and $F$ indicate the size of the time window and feature dimension, respectively, we assume that these time series exhibit periodic behavior. Moreover, it is important to note that the periodicities may vary within the sampled time ranges. To further clarify, let's define $\left[t_1,t_2\right]=\left\{t_1,t_1+1,\ldots,t_2-1,t_2\right\}$. We use the notation $\mathcal{X}_{\left[t_1,t_2\right]} \in \mathbb{R}^{N \times (t_2 - t_1 + 1) \times F}$ to represent the time series sampled from $t_1$ to $t_2$. 

To illustrate, let's consider the scenario where $\mathcal{X}$ represents traffic time series collected from $N$ traffic sensors in a road network. If we sample the data over a period corresponding to a single day for $\mathcal{X}_{\left[t_1,t_2\right]}$, it becomes apparent that the dominant periodicity is approximately 6 hours, as traffic data typically exhibits morning and evening peaks. Conversely, if we sample the data over several days for $\mathcal{X}_{\left[t_1,t_2\right]}$, the prominent period would be one day. Furthermore, it is worth noting that time series can exhibit multiple periodicities. For instance, in the traffic example, there could be periodicities of 6 hours and 1 day.  We introduce the notation ${p}_{\left[t_1,t_2\right]} \in \mathbb{R}$ to denote the prominent periodicity of time series $\mathcal{X}_{\left[t_1,t_2\right]}$.

\noindent {\textbf{Time series representation: }} For a given $\mathcal{X}_{\left[t_1,t_2\right]}$, a representation model $\mathcal{G}\left(\cdot; \mathbf{\theta}\right)$ parameterized by $\mathbf{\theta}$ generates a representation tensor $\mathcal{Y}_{\left[t_1,t_2\right]} = \mathcal{G}\left(\mathcal{X}_{\left[t_1,t_2\right]}; \mathbf{\theta}\right)$. Here, $\mathcal{Y}_{\left[t_1,t_2\right]} \in \mathbb{R}^{N' \times \left(t_2 - t_1 + 1\right) \times F'}$, where $N'$ and $F'$ indicate the dimensions of the modified time series count and the representation feature, respectively. It is important to note that the value of $N'$ varies depending on the choice of $\mathcal{G}$. If we aim to generate an overall representation encompassing all time series, then $N' = 1$. On the other hand, if the goal is to produce a representation for each individual time series, then $N' = N$.

\noindent \textbf{Power Spectral Density:} In signal processing, the power spectral density provides information about the expected signal power at different frequencies of the signal. For example, the periodogram is a measure of spectral density in the Fourier domain. Denoting the discrete Fourier transform as $\mathcal{DFT}\left(\cdot\right)$, the periodogram $\mathbf{\Phi}\left(\cdot\right)$ is computed as:
\begin{equation}
\begin{split}
\mathcal{DFT}\left(w_j\right) = \frac{1}{\sqrt{n}} & \sum_{t=1}^{n} x_t e^{-2\pi i w_j t}, \\
\mathbf{\Phi}\left(w_j\right) = \text{Re}\left(\mathcal{DFT}\left(w_j\right)\right)^2 & + \text{Im}\left(\mathcal{DFT}\left(w_j\right)\right)^2,
\end{split}
\end{equation}
where $x_t$ denotes the time series value at time point $t$, $\text{Re}\left(\cdot\right)$ and $\text{Im}\left(\cdot\right)$ denote the real and imaginary parts, respectively. Each element of the periodogram represents the power at frequency $w_j$, or equivalently, at period $1/w_j$. It is important to note that other transformations, such as discrete cosine transform (DCT) and wavelet transform (DWT), can also be used to calculate the spectral density. If we employ the DCT, the transformation is given by:
\begin{equation}
\begin{split}
\mathcal{DCT}(w_j) =  \left( \frac{n}{2} \right)^{-1/2} \sum^n_{t = 1} & \wedge(t)  x_t \cos\left(\frac{\pi w_j}{2n} \left(2t -1\right) \right), \\
\wedge(t) = & \begin{cases}
\frac{1}{\sqrt{2}}  \quad \text{if} \quad t = 1 \\
1 \quad \text{otherwise}
\end{cases}\\
\mathbf{\Phi}\left(w_j\right) & =  \left|  \mathcal{DCT}\left(w_j\right) \right|.
\end{split}
\end{equation} 

\section{Method}
\label{sec:method}

\begin{figure*}
  \centering
\includegraphics[width=0.9\linewidth]{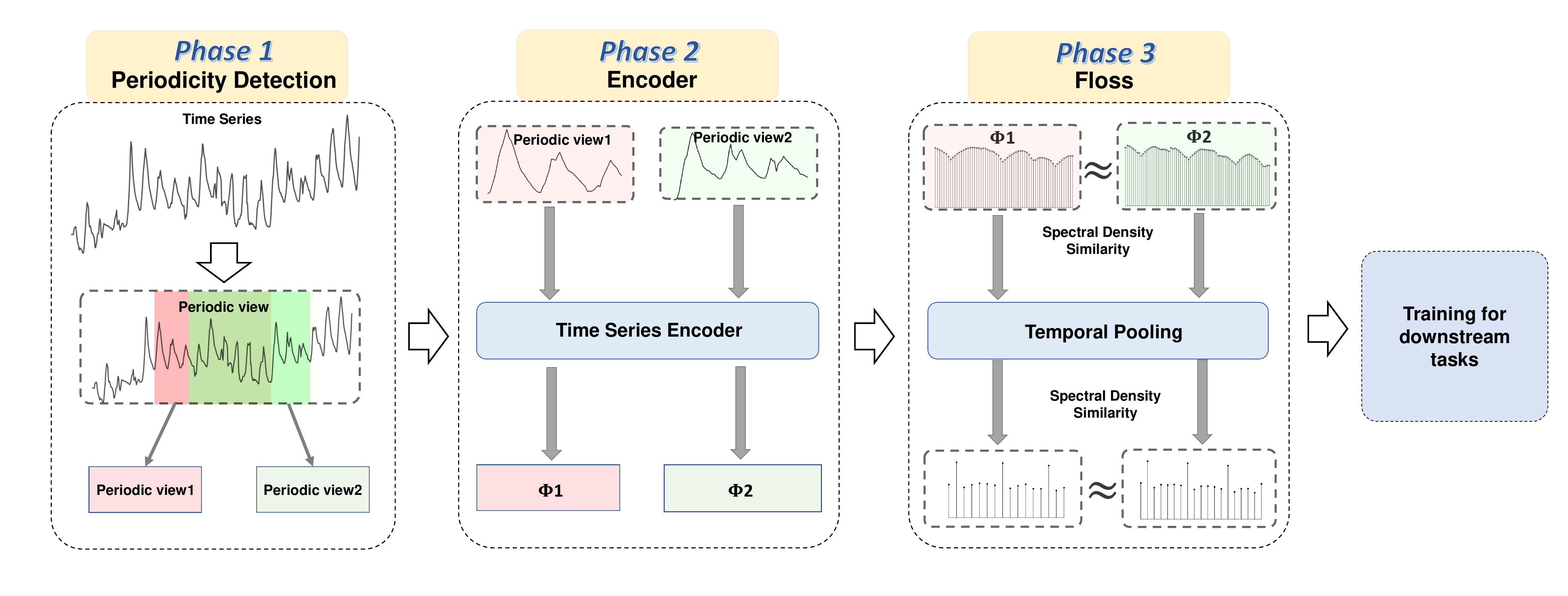}
  \caption{Our framework comprises three critical steps: (1) Periodicity Detection: We automatically detect periodicity patterns from the input time series samples and utilize the detected periodicity to create two views of the input time series. (2) Frequency Domain Similarity Learning: The two periodic views are processed through their respective time series encoders, generating two representations. (3) The Floss algorithm hierarchically calculates the similarities between the spectral densities of the two representations using temporal pooling. The pre-trained encoder can then be directly applied to downstream tasks.}
  \label{fig:framework}
\end{figure*}

In this section, we present the proposed frequency domain loss (Floss) for periodic time series and provide implementation details. Floss is an novel framework that aims to capture the inherent periodic invariance of time series in its learned representations. To accomplish this, the framework incorporates two key steps: a periodicity detection module for generating periodic views and a novel objective that compares the spectral densities of these representations (Figure~\ref{fig:framework}). By doing so, the learned representations are equipped with an awareness of the underlying periodic nature of time series.

\subsection{Periodic Detection and Augmentation}

Assuming the existence of multiple periodicities within each temporal sampled time series $\mathcal{X}_{\left[t_1,t_2\right]}\in \mathbb{R}^{N \times \left(t_2 - t_1 + 1\right) \times F}$, our study focuses on a wide time range $\left[t_1,t_2\right]$ to encompass diverse and significant periodic patterns in the data. In order to create periodic transformation, it is necessary to first identify the underlying periods. This is achieved by calculating the average spectral density using the following procedure:
\begin{equation}
\begin{split}
\mathbf{\hat{\Phi}} = \frac{1}{NF}  \sum^N_{n=1} \sum^F_{f=1} \mathbf{\Phi_{n, f}}, \\
\hat{w} =  \text{arg max}\left(\mathbf{\hat{\Phi}}\right), \\
\hat{p}_{\left[t_1,t_2\right]} = \frac{\left(t_2 - t_1 + 1\right)}{\hat{w}}.
\end{split}
\end{equation}
Here, $\mathbf{\Phi_{n, f}}$ represents the estimated periodogram of the $f$-th feature of the $n$-th time series. The symbol $\mathbf{\hat{\Phi}} \in \mathbb{R}^{t_2 - t_1 + 1}$ denotes the average periodogram across features. It is important to note that the $j$-th value $\mathbf{\Phi}(w_j)$ signifies the intensity of the frequency-$j$ periodic basis function, which is associated with the period length $\frac{(t_2 - t_1 + 1)}{w_j}$. Furthermore, we examine the maximum periodicity $\hat{p}_{\left[t_1,t_2\right]}$ discovered through the periodogram, which corresponds to the highest value observed in $\mathbf{\hat{\Phi}}$.

Although the periodogram is extensively employed for spectral analysis and capturing periodic dynamics, its efficacy can be suboptimal under certain circumstances. Notably, high levels of noise can obfuscate the periodic signals, resulting in inaccurate or potentially deceptive outcomes~\cite{toyoda2013pattern}. Additionally, the periodogram may encounter challenges when faced with complex spectral shapes or irregular patterns, impeding its ability to precisely capture and characterize the underlying periodic dynamics~\cite{vlachos2005periodicity}.

In our approach, we compute a periodogram for each sampled batch, which essentially involves random sampling over the time domain during the training period. We posit that the potential inaccuracies associated with the periodogram can be mitigated by employing this temporal sampling approach. 
By performing random sampling over a wide time range, we increase the number of samples, thereby enhancing the statistical consistency of the estimated periodogram. This approach is supported by empirical validation. For instance, in the field of signal processing, random sampling followed by periodogram analysis has proven effective in identifying periodic signals~\cite{tarczynski2004spectral}. Similarly, in astronomy, this approach has been successfully utilized for periodogram analysis~\cite{vanderplas2015periodograms}.

After obtaining the estimated $\hat{p}_{\left[t_1,t_2\right]}$ for $\mathcal{X}_{\left[t_1,t_2\right]}$, we shift the data along the time axis to exploit the periodic dynamics. We implement this concept through random periodic shifts. In a formal sense, we consider the periodic view of $\mathcal{X}_{\left[t_1,t_2\right]}$ as $\mathcal{X}_{\left[\hat{t}_1,\hat{t}_2\right]}$, where $\hat{t}_1$ and $\hat{t}_2$ are ${t}_1 + a\hat{p}_{\left[t_1,t_2\right]}$ and ${t}_2 + a\hat{p}_{\left[t_1,t_2\right]}$, $a$ is a random integer,

\subsection{Hierarchical Frequency-Domain Loss}
\label{hfl}

Given an encoder $\mathcal{G}(\cdot; \mathbf{\theta})$ parameterized by $\mathbf{\theta}$, along with the original view $\mathcal{X}_{\left[t_1,t_2\right]}$ and its periodic view $\mathcal{X}_{\left[\hat{t}_1,\hat{t}_2\right]}$, our objective is to minimize the difference in power spectral density between the two representations. Let $\mathcal{Y} = \mathcal{G}\left(\mathcal{X}_{\left[t_1,t_2\right]}; \mathbf{\theta}\right)$ and $\hat{\mathcal{Y}} = \mathcal{G}\left(\mathcal{X}_{\left[\hat{t}_1,\hat{t}_2\right]}; \mathbf{\theta}\right)$. Let $\Phi_\mathcal{Y}$ and $\Phi_{\hat{\mathcal{Y}}}$ represent the estimated periodograms of $\mathcal{Y}$ and $\hat{\mathcal{Y}}$ respectively. The loss function for achieving periodic invariance can be defined as follows:
\begin{equation}
\mathcal{L}_f = \frac{1}{N'F'} \| \Phi_\mathcal{Y} - \Phi_{\hat{\mathcal{Y}}} \|_{l1},
\label{equ:loss}
\end{equation}
where $N'$ and $F'$ denote the projected time series and the number of features in $\mathcal{Y}$ and $\hat{\mathcal{Y}}$ respectively.

By minimizing the loss function defined in Equation~\eqref{equ:loss}, we can reap two distinct advantages of preserving periodic invariance. Firstly, it ensures that the representations of the original view and its periodic counterpart exhibit similarity within a specific domain. Secondly, it enables the identification of similar periodic patterns from the representations of both the original view and its periodic view.

However, retaining all frequency components, as in Equation~\eqref{equ:loss}, may lead to subpar representations, as many high-frequency fluctuations in time series can be attributed to noisy inputs. Conversely, exclusively preserving low-frequency components might not be suitable for time series modeling, as certain shifts in trends within the time series carry significant meaning. To better capture information from all frequency components, we propose a hierarchical frequency loss, which compels the encoder to learn representations at multiple scales. Our approach involves hierarchically applying temporal max pooling to the learned features $\mathcal{Y}$ and $\hat{\mathcal{Y}}$, followed by computing their periodic invariance loss. The algorithmic steps for this calculation are outlined in Algorithm~\ref{alg:algorithm1}. Temporal max pooling selects the most prominent element within a given region of the representation, thereby yielding an output that retains the salient features while minimizing noise interference. Furthermore, the temporal pooling operation reduces the temporal dimensionality of the hidden representation. Consequently, the corresponding frequency component of the hidden representation decreases after max pooling, enabling greater emphasis on the low-frequency component. This strategy is reasonable, considering our objective is to encode periodic invariance, which primarily resides within the low-frequency domain.

\begin{algorithm}[tb]
    \caption{Calculating the hierarchical frequency loss}
    \label{alg:algorithm1}
    \begin{flushleft}
    \textbf{Input}:$\mathcal{Y}$, $\hat{\mathcal{Y}}$, a spectral density measure $\mathbf{\Phi}$\\
    \textbf{Parameter}: Pooling scale $\tau$\\
    \textbf{Output}: Hierarchical Loss $\mathcal{L}_{hier}$
    \end{flushleft}
    \begin{algorithmic}[1] 
        \STATE $\mathcal{L}_{hier} \leftarrow \mathcal{L}_f(\mathcal{Y}, \hat{\mathcal{Y}}, \mathbf{\Phi}(\cdot))$;
        \STATE $d \leftarrow 1$
        \WHILE{$\text{length}(\mathcal{Y})> 1$  }
        \STATE $\mathcal{Y} \leftarrow \text{maxpool1d}\left(\mathcal{Y}, \tau \right)$;
        \STATE $\hat{\mathcal{Y}} \leftarrow \text{maxpool1d}\left(\hat{\mathcal{Y}}, \tau \right)$;
        \STATE $\mathcal{L}_{hier} \leftarrow \mathcal{L}_{hier} + \mathcal{L}_f(\mathcal{Y}, \hat{\mathcal{Y}}, \mathbf{\Phi}(\cdot))$;
        \STATE $d \leftarrow d+1$;
        \ENDWHILE
        \STATE $\mathcal{L}_{hier} \leftarrow \mathcal{L}_{hier}/d$;
        \RETURN $\mathcal{L}_{hier}$.
    \end{algorithmic}
\end{algorithm}

In Algorithm~\ref{alg:algorithm1}, the parameter $\tau$ plays a crucial role in controlling the weighting of high-frequency components in the context of max pooling. A larger value of $\tau$ assigns greater importance to the high-frequency parts. For instance, setting $\tau$ to match the temporal length of the feature would effectively equate it to directly comparing the spectral densities of the two features. It is noteworthy that in our experiment, we discovered certain datasets where employing non-hierarchical Floss and directly comparing spectral densities produced superior outcomes. Subsequent analyses will delve deeper into this particular aspect.

\subsection{Training Schemes Under Different Settings}

\begin{figure}
  \centering
  \begin{subfigure}[b]{0.4\textwidth}
    \centering
    \includegraphics[width=\textwidth]{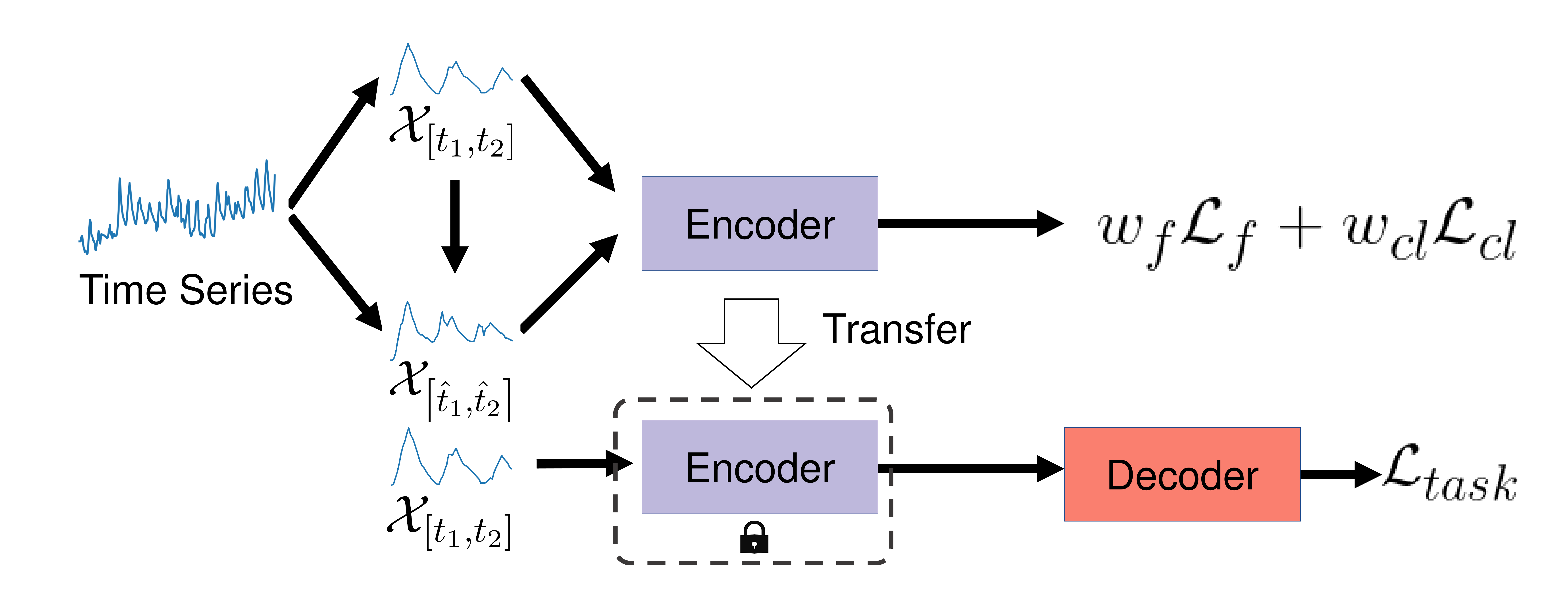}
    \caption{Self-supervised training (fixed encoder parameters after self-supervised training)}
    \label{fig:un-supervised}
  \end{subfigure}
  \hfill
  \begin{subfigure}[b]{0.4\textwidth}
    \centering
    \includegraphics[width=\textwidth]{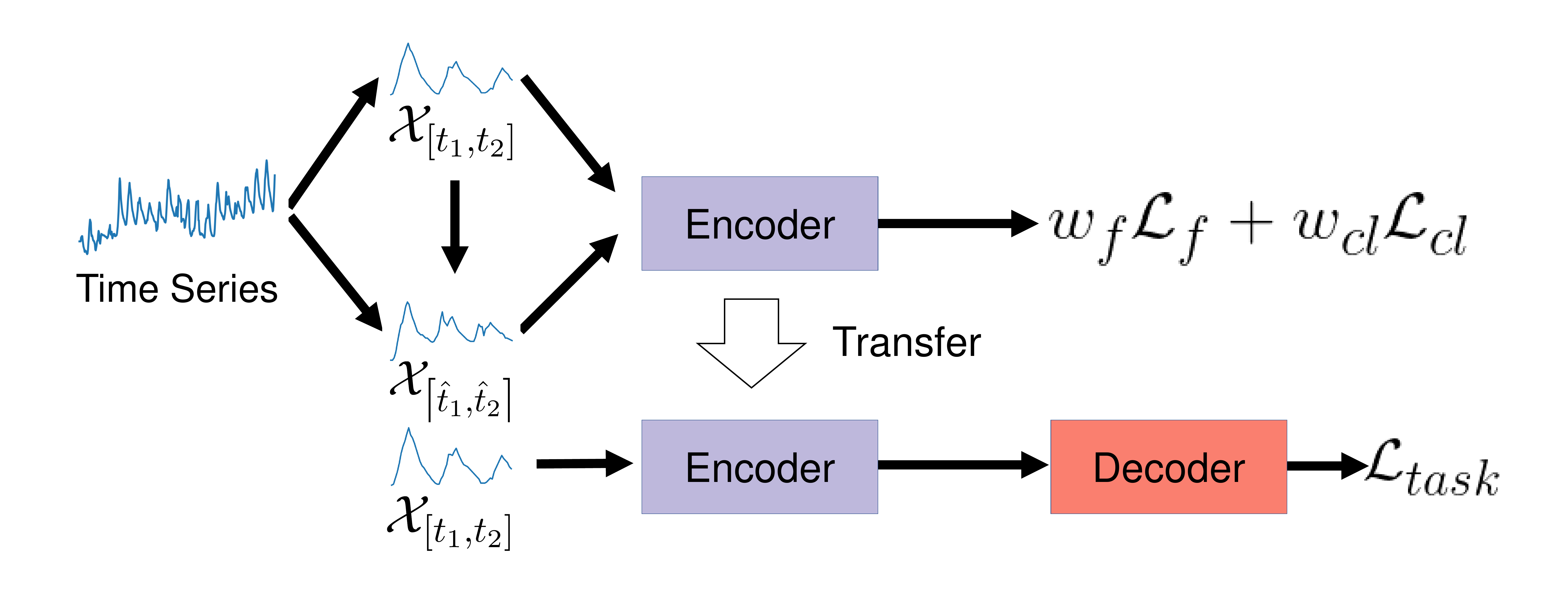}
    \caption{Pre-training then Fine-tuning}
    \label{fig:semi-supervised}
  \end{subfigure}
  \hfill
  \begin{subfigure}[b]{0.4\textwidth}
    \centering
    \includegraphics[width=\textwidth]{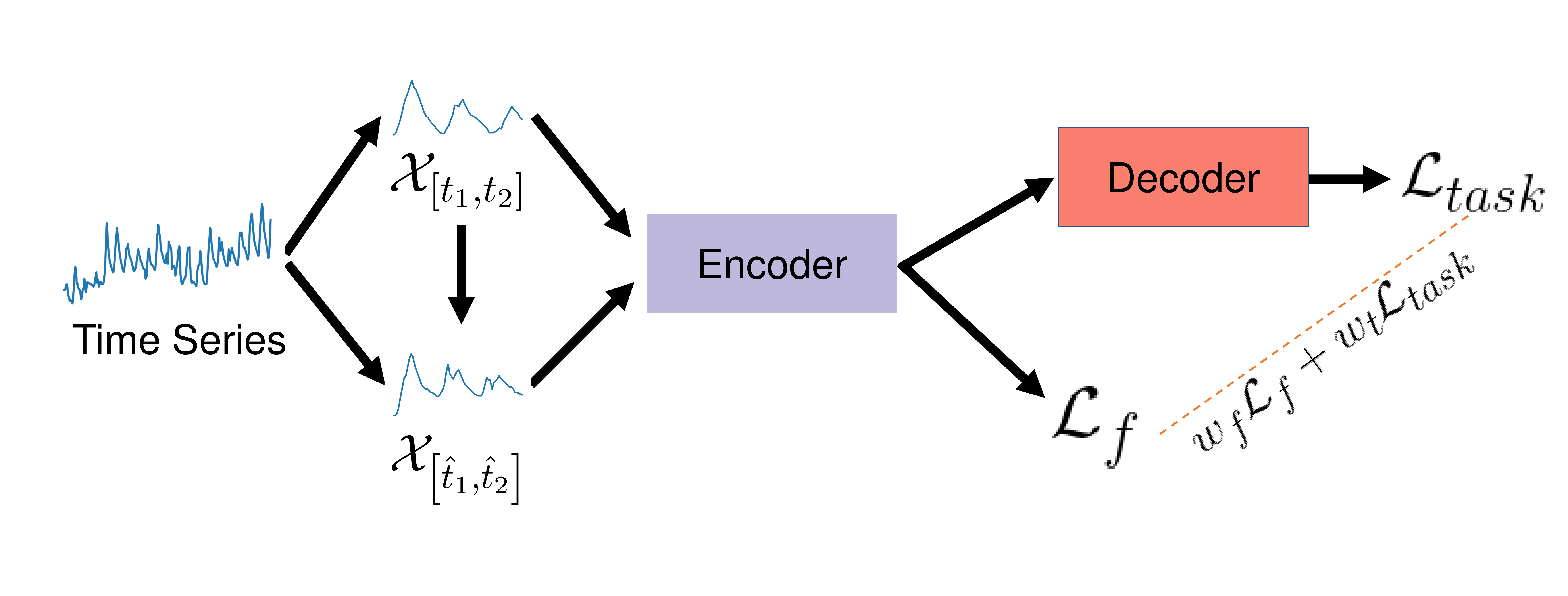}
    \caption{Joint Training}
    \label{fig:supervised}
  \end{subfigure}
  \caption{Illustration of different training schemes.}
  \label{fig:scheme}
\end{figure}

The Frequency-domain loss ({{Floss}}) function, which is proposed in Section~\ref{hfl}, can be readily employed in both supervised and unsupervised learning settings. This section explores the integration of {{Floss}} into unsupervised, semi-supervised, and supervised time series analysis. We summarize the training strategy of different schemes in Figure~\ref{fig:scheme}

\noindent \textbf{1) Self-supervised training:} In the pretraining phase, only the unlabeled time series $\mathcal{X} \in \mathbb{R}^{N \times T \times F}$ are available. First, we randomly sample the original view $\mathcal{X}_{\left[t_1,t_2\right]}$ and its periodic view $\mathcal{X}_{\left[\hat{t}_1,\hat{t}_2\right]}$ from $\mathcal{X}$, considering periodic shifts. To make {{Floss}} compatible with other self-supervised learning schemes, we can apply augmentation techniques such as timestamp masking and random cropping~\cite{yue2022ts2vec} to $\mathcal{X}_{\left[t_1,t_2\right]}$ and $\mathcal{X}_{\left[\hat{t}_1,\hat{t}_2\right]}$. Subsequently, we pass the original and transformed inputs through an encoder $G(\cdot;\mathbf{\theta})$. The {{Floss}} is computed using the representations $G\left(\mathcal{X}_{\left[t_1,t_2\right]};\mathbf{\theta}\right)$ and $G\left(\mathcal{X}_{\left[\hat{t}_1,\hat{t}_2\right]};\mathbf{\theta}\right)$. The {{Floss}} $\mathcal{L}_f$ can be combined with other self-supervised loss functions using a weighted combination $\mathcal{L}_{f}$ and other contrastive learning loss $\mathcal{L}_{cl}$ to train the encoder $G(\cdot;\mathbf{\theta})$. During this stage, the downstream tasks are assumed to be unknown. Finally, we follow the same protocol as \cite{franceschi2019unsupervised}, where a decoder is trained on top of the representations $G\left(\mathcal{X}_{\left[\hat{t}_1,\hat{t}_2\right]};\mathbf{\theta}\right)$ to handle the downstream tasks. It is important to note that the parameters $\mathbf{\theta}$ of the encoder remain fixed during the final training phase.

\noindent \textbf{2) Pre-training then Fine-tuning:} The procedure for pretraining in the semi-supervised setting is similar to that of the unsupervised setting. However, during the fine-tuning stage, the optimized model parameters $\mathbf{\theta}$ of $G(\cdot;\mathbf{\theta})$ are further fine-tuned to transition from $G(\cdot;\mathbf{\theta})$ to $G(\cdot;\mathbf{\phi})$ using the downstream tasks.

\noindent \textbf{3) Joint training under supervised learning setting:} In the joint training approach, both the encoder and decoder are trained simultaneously. In this scenario, the {{Floss}} serves as an auxiliary regularization term during training, providing additional self-supervision signals that contribute to enhancing generalization. Specifically, in this setting, both the unlabeled time series $\mathcal{X} \in \mathbb{R}^{N \times T \times F}$ and their corresponding labels $\mathcal{D}$ are available during the training phase. The encoder $G(\cdot;\mathbf{\theta})$ is trained using a weighted combination of the {{Floss}} $\mathcal{L}_{f}$ and the supervised loss $\mathcal{L}_{task}$.

\section{Experiments}
\label{sec:exp}

In this section, we assess the effectiveness of {Floss} in periodic time series forecasting, classification, and anomaly detection. Our primary objective in this study is to determine whether incorporating {Floss} can enhance the performance of current supervised and unsupervised representation learning frameworks.

\subsection{Multivariate time series forecasting}

\subsubsection{ Existing Algorithms.} We consider three representative multivariate time series forecasting models: 1). TS2Vec~\cite{yue2022ts2vec}: This is a purely unsupervised learning model. TS2Vec employs contrastive learning in a hierarchical manner on augmented views. Its encoder is based on a lightweight temporal convolutional network. After training, the encoder remains fixed, and ridge regression is used for the forecasting task. To integrate Floss into TS2Vec, we adapt the augmentation strategy of TS2Vec to incorporate periodic shifts. We randomly sample two segments $[t_1 - j_1, t_2 + k_1]$ and $[t_1 + a\hat{p}{\left[t_1,t_2\right]} - j_2, t_2 + a\hat{p}{\left[t_1,t_2\right]} + k_2]$, where $a, j_1, j_2, k_1,$ and $k_2$ are random integers. We also apply TS2Vec's timestamp mask strategy to the time series segments. Then, we train our model using a weighted sum of the frequency loss and contrastive loss from TS2Vec on the representations of the segments $[t_1, t_2]$ and $[t_1 + a\hat{p}{\left[t_1,t_2\right]}, t_2 + a\hat{p}{\left[t_1,t_2\right]}]$. The estimated periodicity and frequency loss are computed using discrete cosine transformation (DCT). 2) PatchTST~\cite{nie2023a}: This model employs a vision Transformer-style architecture for multivariate time series forecasting and utilizes pre-training and fine-tuning techniques for training. In the self-learning phase, the model is trained to reconstruct masked time series patches. After self-training, the transformer is fine-tuned for downstream multivariate forecasting tasks. In our approach, Floss collaborates with the reconstruction loss in a weighted sum fashion during the self-training phase. 3) Informer~\cite{zhou2021informer}: This transformer model is a milestone in time series forecasting and is trained using a purely supervised learning approach. We incorporate Floss to regularize its hidden representation, specifically the layer before the final layer. The model is trained by combining the forecasting loss and the proposed frequency loss using a weighted sum. 

Not only do we choose models based on the paradigms of different training schemes, but the sizes of these three models are also representative. TS2Vec has a relatively small structure, Informer is of medium size, while PatchTST is a larger model.

\subsubsection{Public Datasets.} We assess the effectiveness of our proposed Floss by evaluating its performance on 8 widely-used datasets, namely Weather, Exchange, Electricity, ILI, and 4 ETT datasets (ETTh1, ETTh2, ETTm1, ETTm2). These datasets are commonly employed for benchmarking purposes and are publicly available on~\cite{Zeng2022AreTE}. For the TS2Vec model, we allocated 60\% of the data for training, 20\% for validation, and 20\% for testing. For the PatchTST and Informer models, we allocated 70\% of the data for training, 10\% for validation, and 20\% for testing. The statistics of those datasets are summarized in Table~\ref{tab:data}.

\begin{table*}[t]
  \caption{Statistics of popular datasets for benchmark.}
  \label{tab:data}
  \begin{tabular}{ccccccc}
    \toprule
Datasets & ETTh1\&ETTh2 & ETTm1 \&ETTm2 & Electricity & Exchange-Rate & Weather & IL \\
\midrule
Variates & 7& 7 & 321 & 8 & 21 & 7 \\
Timesteps & 17,420 & 69,680 & 26,304 & 7,588 & 52,696 & 966 \\
Granularity &1hour & 15min &  1hour & 1day & 10min & 1week \\
    \bottomrule
  \end{tabular}
\end{table*}

\subsubsection{Experimental Settings.} Following previous works~\cite{zhou2022fedformer, Zeng2022AreTE, zhou2021informer}, we use Mean Squared Error (MSE) and Mean Absolute Error (MAE) as the core metrics to compare performance. All of the models follow the same experimental setup with a prediction length of $T \in \{24, 36, 48, 60\}$ for the ILI dataset and $T \in \{96, 192, 336, 720\}$ for other datasets, as mentioned in the original papers. For PatchTST and Informer, the lookback window is set to $L = 96$. We adhere to the standard protocol and split all datasets into training, validation, and test sets in chronological order using a ratio of 7:1:2. For TS2Vec, the lookback window is set equal to the prediction length $T$, and all datasets are split into training, validation, and test sets in the ratio of 6:2:2 (same as the original paper~\cite{yue2022ts2vec}). 

The detailed hyper-parameter configurations of informer-Floss are set as follows: The batch size for all datasets is set to 32. Loss weights for different datasets are as follows: Weather (original forecasting loss weight= 0.3, Floss weight= 2), Exchange (original loss weight= 0.3, Floss weight= 0.7 for 96-step ahead prediction, Floss weight= 0.8 for all other prediction horizons), Electricity (original loss weight= 0.3, Floss weight= 2), ILI (original loss weight= 0.3, Floss weight= 0.5), ETTh1 (original loss weight= 0.3, Floss weight= 1), ETTh2 (original loss weight= 0.5, Floss weight= 8), ETTm1, and ETTm2 (original loss weight= 0.5, Floss weight= 8).

The detailed hyper-parameter configurations of TS2Vec-Floss are as follows: The batch size is set to 16, the Floss weight for the contrastive loss of TS2Vec is set to 1, and the loss for the contrastive loss of TS2Vec is assigned a value of 1. Similarly, the detailed hyper-parameter configurations of PatchTST-Floss are as follows: During pretraining, the reconstruction loss is set to 0.3, the Floss weight is set to 1, and the mask ratio is set to 0.4. Additionally, the batch size for Weather, Electricity, ETTh1, ETTh2, ETTm1, and ETTm2 datasets is set to 8, while for Exchange and ILI datasets, it is set to 16.

\begin{table*}[!t]
    \centering
        \caption{Errors of Multivariate Time Series Forecasting. The improved results are in bold.}
    \begin{adjustbox}{center}
    \begin{tabular}{llllllllllllll}
    \toprule   
         \multicolumn{2}{c}{Dataset} & \multicolumn{2}{c}{Informer} & \multicolumn{2}{c}{Informer-Floss}&\multicolumn{2}{c}{TS2vec}&\multicolumn{2}{c}{TS2vec-Floss} &\multicolumn{2}{c}{PatchTST} &\multicolumn{2}{c}{PatchTST-Floss} \\
         \multicolumn{2}{c}{Metric} & MSE & MAE & MSE & MAE& MSE & MAE& MSE & MAE& MSE & MAE& MSE & MAE \\
        \midrule 
         \multirow{4}{*}{Weather} & 96 & 0.427 & 0.460 & \textbf{0.277} & \textbf{0.370} & 1.719 & 0.921 & \textbf{1.278} & \textbf{0.840} & 0.144 & 0.192 & \textbf{0.125} & \textbf{0.173} \\
        ~ & 192 & 0.346 & 0.414 & 0.361 & \textbf{0.402} & 1.650 & 0.925 & \textbf{1.360} & \textbf{0.882} & 0.191 & 0.241 & \textbf{0.183} & \textbf{0.229} \\
        ~ & 336 & 0.583 & 0.543 & \textbf{0.407} & \textbf{0.408} & 1.949 & 1.043 & \textbf{1.318} & \textbf{0.876} & 0.244 & 0.280 & \textbf{0.232} & \textbf{0.271} \\
        ~  & 720 & 0.916 & 0.705 & \textbf{0.837} & \textbf{0.668} & 2.718 & 1.287 & \textbf{1.559} & \textbf{0.972} & 0.314 & 0.331 & \textbf{0.301} & \textbf{0.325} \\
        \midrule 
        \multirow{4}{*}{Exchange}&96 & 0.841 & 0.746 & \textbf{0.753} & \textbf{0.705} & 0.498 & 0.527 & \textbf{0.422} & \textbf{0.484} & 0.099 & 0.224 & 0.099 & 0.225 \\
        ~ & 192& 1.132 & 0.847 & \textbf{1.180} & 0.859 & 1.112 & 0.781 & \textbf{0.851} & \textbf{0.687} & 0.210 & 0.331&0.210 &\textbf{0.330} \\
        ~ & 336 & 1.475 & 0.956 & 1.510 & 0.974 & 1.561 & 0.967 & 1.571 & \textbf{0.944} & 0.404 & 0.468&0.424&0.478 \\
        ~  & 720 & 2.548 & 1.328 & 2.606 & 1.362 & 2.688 & 1.266 & \textbf{1.860} & \textbf{1.052} & 1.039 & 0.769 & \textbf{0.902} & \textbf{0.720} \\ 
        \midrule 
        \multirow{4}{*}{Electricity}&96 & 0.304 & 0.393 & \textbf{0.285} &\textbf{0.380} & 0.452 & 0.492 & \textbf{0.422} & \textbf{0.463} & 0.135 & 0.231 & \textbf{0.129} & \textbf{0.228} \\
        ~ & 192& 0.327 & 0.417 & \textbf{0.297} & \textbf{0.390} & 0.461 & 0.498 & \textbf{0.423} & \textbf{0.465} & 0.150 & 0.244 & \textbf{0.149} & \textbf{0.242} \\
        ~ & 336 & 0.333 & 0.422 & \textbf{0.302} & \textbf{0.396} & 0.472 & 0.491 & \textbf{0.426} & \textbf{0.468} & 0.165 & 0.259 & \textbf{0.159} & 0.260 \\
        ~ & 720 & 0.351 & 0.427 & \textbf{0.325} & \textbf{0.406} & 0.544 & 0.547 & \textbf{0.513} & \textbf{0.516} & 0.203 & 0.292 & \textbf{0.201} & \textbf{0.287} \\  
        \midrule 
       \multirow{4}{*}{ILI} &24 & 5.940 & 1.720 & \textbf{5.460} & \textbf{1.580} & 3.349 & 1.168 & 3.686 & 1.276 & 2.883 & 1.189 & 2.962 & 1.200 \\
        ~ & 36 & 4.999 & 1.508 & 5.300 & 1.541 & 3.671 & 1.244 & 4.131 & 1.399 & 2.986 & 1.195 & \textbf{2.850} & \textbf{1.169} \\
        ~ &48  & 5.004 & 1.542 & 5.319 & 1.570 & 4.150 & 1.324 & 4.153 & 1.364 & 3.411 & 1.287 & \textbf{2.899} &\textbf{1.174} \\
        ~  & 60  & 5.403 & 1.554 & 5.631 & 1.589 & 4.231 & 1.340 & \textbf{4.185} & 1.359 & 3.207 & 1.233 & \textbf{3.142} & \textbf{1.227} \\ 
        \midrule 
        
       \multirow{4}{*}{ETTh1} & 96 & 0.941 & 0.769 & \textbf{0.801} & \textbf{0.695} & 0.699 & 0.592 & 0.804 & 0.666 & 0.373 & 0.402 & \textbf{0.368} & \textbf{0.397} \\
        ~ & 192 & 1.007 & 0.786 & \textbf{0.867} & \textbf{0.713} & 0.789 & 0.643 & 0.876 & 0.704 & 0.403 & 0.419 & 0.403 & 0.421 \\
        ~ & 336 & 1.038 & 0.784 & 1.140 & 0.859 & 0.907 & 0.709 & 0.969 & 0.750 & 0.443 & 0.449 & \textbf{0.432} & \textbf{0.441} \\
        ~  & 720  & 1.144 & 0.857 & 1.184 & 0.883 & 1.084 & 0.800 & \textbf{0.969} & \textbf{0.750} & 0.482 & 0.490 & \textbf{0.451} & \textbf{0.472} \\ 
        \midrule 
           
       \multirow{4}{*}{ETTh2}& 96 & 3.283 & 1.502 & \textbf{2.763} &\textbf{ 1.372} & 1.034 & 0.806 & 1.065 & 0.808 & 0.287 & 0.344 & \textbf{0.285} & \textbf{0.342} \\
        ~ & 192& 4.371 & 1.815 & \textbf{4.110} & \textbf{1.713} & 1.973 & 1.118 & 2.177 & 1.163 & 0.363 & 0.392 & \textbf{0.359} & \textbf{0.386}\\
        ~ & 336 & 4.215 & 1.642 & \textbf{3.910} & 1.656 & 2.831 & 1.319 & \textbf{2.398} & \textbf{1.238} & 0.375 & 0.409 & 0.376 & \textbf{0.405} \\
        ~  & 720 & 3.656 & 1.619 & \textbf{3.222} & \textbf{1.541} & 2.561 & 1.353 & 2.578 & \textbf{1.331} & 0.411 & 0.443 & \textbf{0.399} & \textbf{0.428} \\ 
        \midrule 
           
       \multirow{4}{*}{ETTm1}& 96 & 0.657 & 0.575 & \textbf{0.629} & 0.582 & 0.611 & 0.551 & \textbf{0.565} & \textbf{0.519} & 0.282 & 0.339 & \textbf{0.281} & \textbf{0.328} \\
        ~ & 192& 0.725 & 0.619 & 0.744 & 0.647 & 0.675 & 0.589 & \textbf{0.616} & \textbf{0.553} & 0.329 & 0.369 & \textbf{0.319} & \textbf{0.356}\\
        ~ & 336 & 0.725 & 0.619 & 1.053 & 0.819 & 0.725 & 0.621 & \textbf{0.681} & \textbf{0.593} & 0.358 & 0.387 & \textbf{0.349} & \textbf{0.378} \\
        ~  & 720 & 1.133 & 0.845 & \textbf{0.997} & \textbf{0.778} & 0.810 & 0.671 & \textbf{0.763} & \textbf{0.643} & 0.411 & 0.415 & \textbf{0.397} & \textbf{0.411} \\ 
        \midrule 
       
       \multirow{4}{*}{ETTm2}&96 & 0.555 & 0.462 & \textbf{0.488} & 0.514 & 0.443 & 0.495 & \textbf{0.371} & \textbf{0.447} & 0.164 & 0.254 & \textbf{0.158} & \textbf{0.233} \\
        ~ & 192& 0.695 & 0.686 & 0.715 & \textbf{0.652} & 0.615 & 0.598 & \textbf{0.546} & \textbf{0.561} & 0.220 & {0.294} & \textbf{0.197} & \textbf{0.252} \\
        ~ & 336 & 1.270 & 0.871 & \textbf{1.119} & \textbf{0.805} & 0.975 & 0.765 & \textbf{0.863} & \textbf{0.721} & 0.271 & 0.327 & \textbf{0.248} & \textbf{0.319} \\
        ~  & 720 & 3.171 & 1.367 & 3.414 & 1.374 & 2.024 & 1.093 & \textbf{1.977} & 1.104 & 0.354 & 0.381 & \textbf{0.339} & \textbf{0.355} \\ 
        \midrule 
        Avg. & ~ & 1.868 & 0.935 & 1.812 & 0.912  & 1.562 & 0.860 & 1.449 & 0.831 & 0.666 & 0.465 & 0.635 & 0.452 \\ 

       Improvements. & ~ &  &  & {\color{red}$\downarrow 3.0\%$} &   {\color{red}$\downarrow 2.4\%$} &  & & {\color{red}$\downarrow 7.2\%$} & {\color{red}$\downarrow 3.4\%$}  &  &  & {\color{red}$\downarrow 4.6\%$}  & {\color{red}$\downarrow 2.8\%$} \\ 
    \bottomrule 
    \end{tabular}
    \end{adjustbox}
    \label{tab:multi_pre}
\end{table*}

\subsubsection{Experimental Results} Table~\ref{tab:multi_pre} shows the multivariate long-term forecasting results. It should be noted that we reran the experiments for fair comparison; therefore, the performance of Informer, TS2Vec, and PatchTST is slightly better than what was reported in the original literature. We use bold text to highlight the improved performance and red color to indicate the average improvements. The key observations are as follows:

First, the inclusion of Floss enhances the overall performance of all three representative models. This demonstrates that Floss effectively utilizes informative features within the frequency domain, leading to improved forecasting performance.

Secondly, Floss performs remarkably well on the Electricity dataset, which includes the largest number (321) of time series in our experiments. Improvements are observed in all cases, indicating that Floss has the ability to encode shared frequency information from a large number of time series, thereby enhancing forecasting performance.

Thirdly, the inclusion of Floss does not consistently outperform the models without it. This could be attributed to the random factors involved in the training process with Floss, such as the random sampling for periodicity detection and the random shift using the detected periodicity. These factors might prevent the models from consistently leveraging valuable information. Future studies should address this issue to ensure more consistent results.

\begin{figure*}
  \centering
  \begin{subfigure}[b]{0.315\textwidth}
    \centering
    \includegraphics[width=\textwidth]{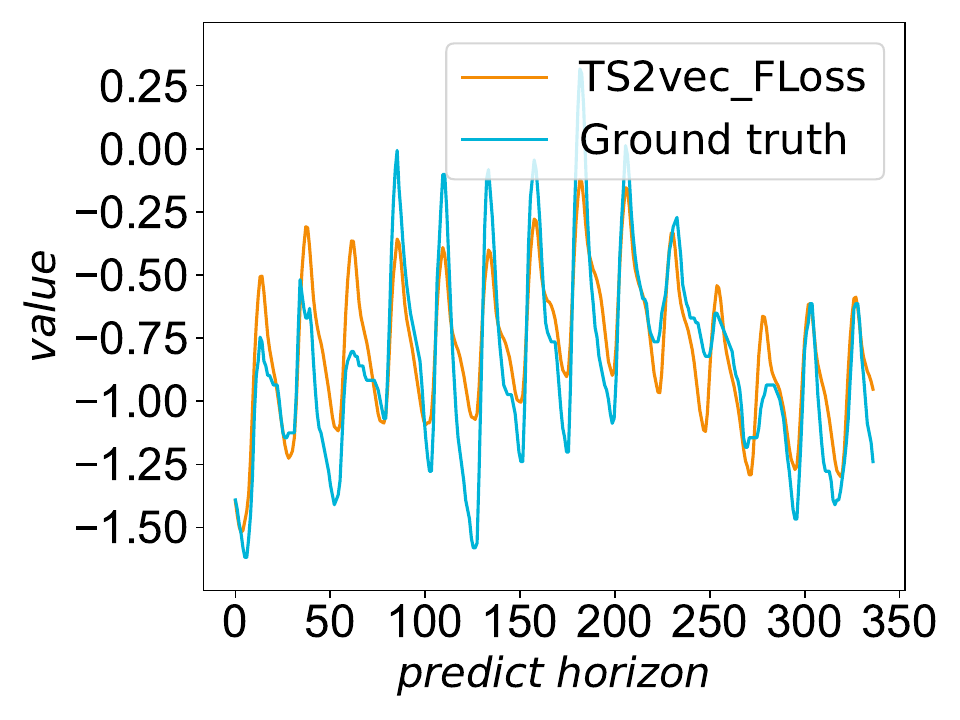}
    \caption{TS2Vec-Floss forecasting results on \textit{ETTh2}}
    \label{fig:fore1}
    \vspace{\baselineskip} 
  \end{subfigure}
  \hfill
   \begin{subfigure}[b]{0.315\textwidth}
    \centering
    \includegraphics[width=\textwidth]{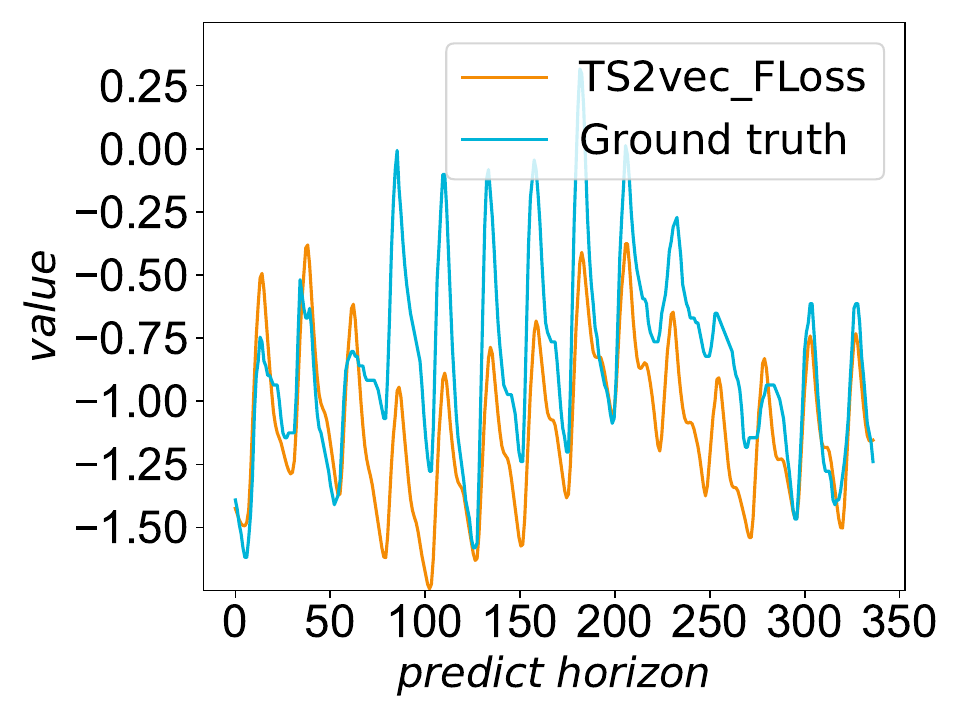}
    \caption{TS2Vec forecasting results on \textit{ETTh2}}
    \label{fig:fore2}
    \vspace{\baselineskip} 
  \end{subfigure}
  \hfill
  \begin{subfigure}[b]{0.315\textwidth}
    \adjincludegraphics[width=\textwidth, valign=t]{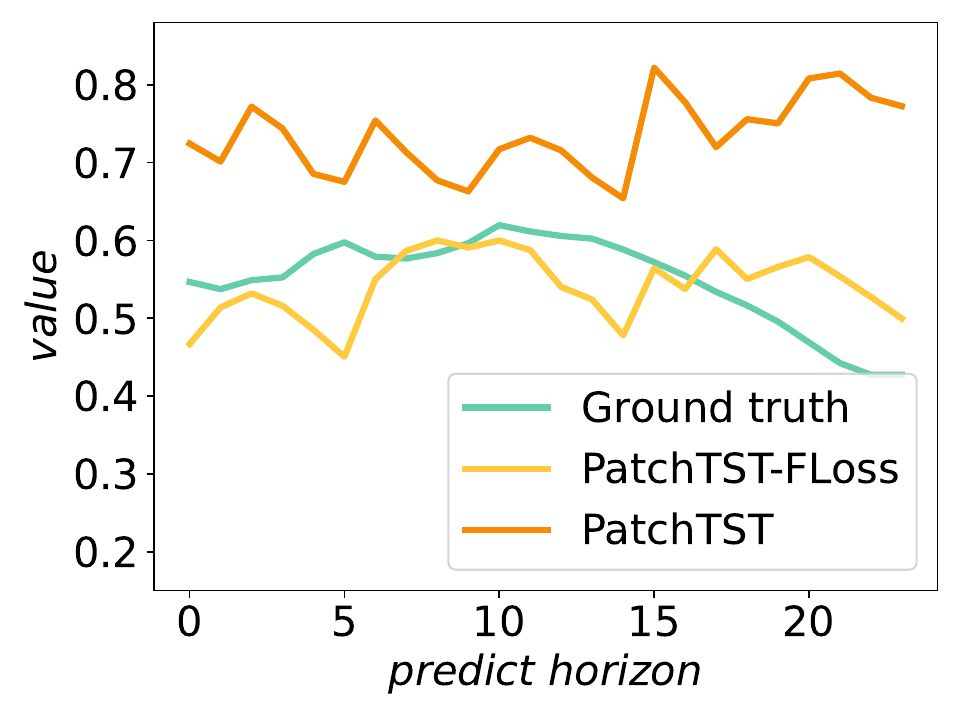}
    \caption{PatchTST and PatchTST-Floss forecasting results on \textit{weather}}
    \label{fig:fore3}
  \end{subfigure}
  \caption{Illustration of the long-term forecasting output of model w/o Floss on \textit{ETTm2} and \textit{weather} datasets (Y-axis: forecasting horizon).}
  \label{fig:forecast}
\end{figure*}

As depicted in Figure~\ref{fig:forecast}, the prediction results of PatchTST and TS2Vec w/o Floss are presented for the \textit{ETTh2} and \textit{weather} datasets. In the long-term forecasting horizon of \textit{ETTh2}, Floss demonstrates its superiority in handling distribution shifts and trend-seasonality features in comparison to TS2Vec. This advantage can be attributed to the enhanced ability of Floss to effectively leverage trend information by regularizing representations in the frequency domain. Figure~\ref{fig:fore3} further demonstrates the superior performance of PatchTST-Floss in both short-term and long-term forecasting tasks, highlighting the significant benefits introduced by Floss in the context of forecasting.

\subsection{Unsupervised Time Series Classification with TS2Vec}

\subsubsection{Experimental Setup.}  In this section, we combine {Floss} with the state-of-the-art (SOTA) unsupervised framework TS2Vec~\cite{yue2022ts2vec}, which has outperformed several supervised learning frameworks. We utilize the same convolutional encoder as described in \cite{yue2022ts2vec}. Additionally, we modify the sampling strategy of TS2Vec to create periodic shifts, aligning it with the settings used for the aforementioned multivariate time series forecasting. Following the pre-training phase, we train an SVM classifier with an RBF kernel on top of the instance-level representations to perform predictions.

\subsubsection{Public Datasets.}  We evaluate the effectiveness of our proposed Floss by assessing its classification performance on two widely-used datasets: the UCR archive~\cite{dau2019ucr} and UEA archive~\cite{bagnall2018uea}. The UCR archive consists of 128 univariate datasets, while the UEA archive contains 30 multivariate datasets. For each dataset considered, we utilize its original train/test split. We conduct unsupervised training of an encoder using the train set of each dataset. Subsequently, we train an SVM classifier with a RBF kernel on top of the learned features, utilizing the train labels of the dataset. Finally, we output the corresponding classification score on the test set. For the hyperparameter settings, batch size is 16, the contrastive loss weight is 1, Floss weight is 1.

\subsubsection{Compared Baselines.} We perform comprehensive experiments on time series classification to assess the classification performance of our approach, in comparison to other unsupervised time series representation models, namely T-Loss~\cite{franceschi2019unsupervised}, TS-TCC~\cite{ijcai2021-324}, TST~\cite{10.1145/3447548.3467401}, and TNC~\cite{tonekaboni2020unsupervised}. Additionally, we include DTW (Dynamic Time Warping)~\cite{muller2007dynamic} as a baseline, employing a one-nearest-neighbor classifier with DTW as the distance measure.

\begin{table}[h]
\centering
\caption{Time series classification results compared to other time series representation methods on 125 UCR datasets and 29 UEA datasets.}
\begin{adjustbox}{center}
\begin{tabular}{lcc}
\midrule 
{Method}  & {125 UCR datasets } & {29 UEA datasets} \\
\midrule 
DTW & 0.727 & 0.650\\
TNC & 0.761 & 0.677\\
TST & 0.641 & 0.635\\
TS-TCC & 0.757 & 0.682\\
T-Loss & 0.806 & 0.675\\
TS2Vec & 0.830 & 0.712\\
{TS2Vec-Floss} & \textbf{0.849} & \textbf{0.739} \\
\midrule 
\end{tabular} 
\end{adjustbox}
\label{tab:classification}
\end{table}

\begin{figure}
  \centering
  \includegraphics[width=\linewidth]{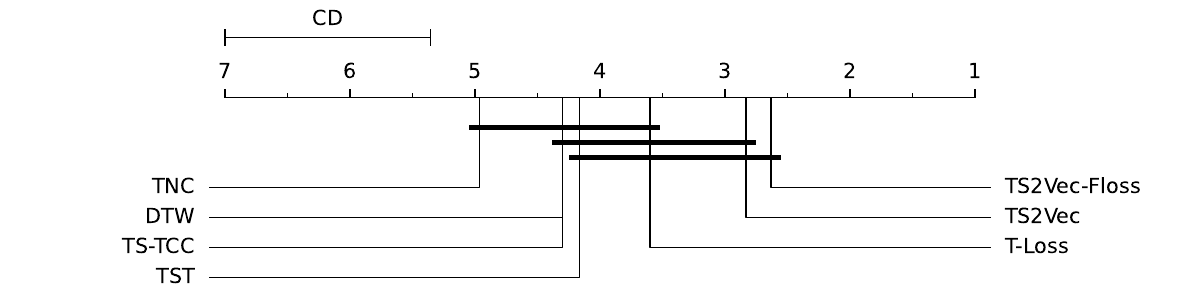}
  \caption{Critical Difference (CD) diagram of representation
learning methods on time series classification tasks with a
confidence level of 95\%.}
  \label{fig:class_cd}
  
\end{figure}

\subsubsection{Experimental Results.} The evaluation results are summarized in Table~\ref{tab:classification}. Floss demonstrates a significant improvement compared to other representation learning methods on both the UCR and UEA datasets. Specifically, Floss achieves an average increase of 2.3\% in classification accuracy over TS2Vec across 125 UCR datasets and 3.0\% across 29 UEA datasets. It is important to note that the periodicity detection module is applicable to all UCR and UEA datasets, and comprehensive results of TS2Vec-Floss on all datasets can be found in the supplementary materials. Critical Difference diagram~\cite{demvsar2006statistical} for Nemenyi tests on all datasets (including 125 UCR and 29 UEA datasets) is presented in Figure~\ref{fig:class_cd}, where classifiers that are not connected by a bold line are significantly different in average rank. Unlike existing baselines that neglect periodic information, Floss utilizes hierarchical frequency domain comparison between different periodic views, resulting in enhanced performance.

\subsection{Unsupervised Time Series Classification with TS-TCC}

\subsubsection{Experimental Setup.} We combine {Floss} with another representative model for time series representation called TS-TCC~\cite{ijcai2021-324}. To evaluate our model, we conduct human activity recognition, sleep stage classification, and epileptic seizure prediction tasks using open-source datasets. Following the approach of TS-TCC, we perform pre-training and downstream task fine-tuning for 40 epochs. During the pre-training phase, we incorporate Floss with the contrastive loss function of TS-TCC. In contrast to the TS2Vec setup, we introduce a separate periodic augmentation alongside the jitter and scale augmentation of TS-TCC. Moreover, Floss is computed based solely on the original and periodic views of the time series data. 
The encoder is trained using Adam with a weighted sum of Floss and the original loss of TS-TCC. We maintain the same hyperparameters as those reported in \cite{ijcai2021-324}. For the loss weights, the original loss weight is 0.3 and Floss weight is 2.

\subsubsection{Public Datasets.}  We assess the classification performance of our proposed Floss by evaluating it on three widely-used datasets: 1. UCI HAR dataset~\cite{anguita2013public}: This dataset contains sensor readings for 30 subjects performing 6 activities. The sample rate of the HAR dataset is 60Hz. 2.Sleep-EDF~\cite{goldberger2000physiobank}: This dataset includes whole-night PSG sleep recordings, with a sampling rate of 100Hz. 3. The Epileptic Seizure Recognition dataset~\cite{andrzejak2001indications}: This dataset consists of EEG recordings from 500 subjects, where the brain activity was recorded for each subject for 23.6 seconds. We split the data into 60\%, 20\%, and 20\% for training, validation, and testing, respectively. For the Sleep-EDF dataset, we perform a subject-wise split to prevent overfitting. We repeat the experiments five times using five different seeds. During the fine-tuning phase, we train a linear classifier (a single MLP layer) on top of a frozen self-supervised pretrained encoder model to perform classification. 

\begin{table*}[h]
\centering
\caption{Time series classification results compared to other time series representation methods on HAR, Sleep-EDF and Epilepsy.}
\begin{adjustbox}{center}
\begin{tabular}{ccccccc}
 \toprule  
{Datasets} & \multicolumn{2}{c}{HAR} & \multicolumn{2}{c}{Sleep-EDF} & \multicolumn{2}{c}{Epilepsy}\\
\midrule 
{Metric} & ACC & MF1 & ACC & MF1 & ACC & MF1  \\
\midrule 
{CPC} & {$83.85 \pm 1.51$} & {$83.27 \pm 1.66$} &{$82.82 \pm 1.68$}& {$\textbf{73.94} \pm \textbf{1.75}$} & {$96.61 \pm 0.43$} & {$94.44 \pm 0.69$}\\
{SimCLR}&  {$80.97 \pm 2.46$} & {$80.19 \pm 2.64$} &{$78.91 \pm 3.11$}& {$68.60 \pm 2.71$} & {$96.05 \pm 0.34$} & {$93.53 \pm 0.63$} \\
{TS-TCC}& {$90.37 \pm 0.34$} & {$90.38 \pm 0.39$} &{$83.00 \pm 0.71$}& {$73.57 \pm 0.74$} & {$97.23 \pm 0.10$} & {$95.54 \pm 0.08$} \\
{TS-TCC-{Floss}
}& {$\textbf{90.86} \pm \textbf{0.34}$} & {$\textbf{90.56} \pm \textbf{0.35}$} & {$\textbf{83.70} \pm \textbf{0.45}$}& {$73.53 \pm 0.39$} & {$\textbf{97.41} \pm \textbf{0.17}$} & {$\textbf{97.75} \pm \textbf{0.00}$}\\
\midrule 
\end{tabular}
\label{tab:TS-TCC}
\end{adjustbox}
\end{table*}

\subsubsection{Experimental Results.}  We report the accuracy (ACC) and macro F1 score (MF1) of the TS-TCC-Floss, raw TS-TCC~\cite{ijcai2021-324}, CPC~\cite{oord2018representation} and
SimCLR~\cite{chen2021exploring} in Table~\ref{tab:TS-TCC}. Similar to the findings observed in the TS2Vec experiments, the integration of Floss yields significant enhancements in the performance of TS-TCC. Notably, an intriguing aspect emerges when examining the three datasets employed in this study, wherein the sampling periods are comparatively short. Intuitively, discerning the presence of short-term periodic information in these datasets poses a formidable challenge. However, employing Floss still yields notable improvements across these datasets. This phenomenon can be attributed to the inherent capacity of Floss to autonomously detect periodicity, thereby effectively capturing imperceptible quasi-periodic variations within the data. Consequently, Floss exhibits an automatic mechanism for augmenting the representational quality of existing models, thereby advancing their efficacy.

\subsection{Unsupervised Anomaly Detection}

\subsubsection{Experimental Setup} For anomaly detection, we follow the streaming evaluation protocol, where the task is to determine whether the last point $t$ is an anomaly. As same as in \cite{yue2022ts2vec}, we define the anomaly score as the dissimilarity between the representations computed from the original series and the one with a mask at the last time point. We use the same computation strategy as described in \cite{yue2022ts2vec} to compute anomalies. Two public datasets are used to evaluate our model. \textit{Yahoo}\footnote{\url{https://yahooresearch.tumblr.com/post/114590420346/a-benchmark-dataset-for-time-series-anomaly}} is a benchmark dataset for anomaly detection, which includes 367 hourly sampled time series with tagged anomaly points. \textit{KPI}~\cite{ren2019time} includes multiple minutely sampled real KPI curves from various Internet companies. In the normal setting, each time series sample is split into two halves according to the time order, where the first half is used for unsupervised training and the second half is used for evaluation. We also evaluate the cold-start problem, in which the TS2Vec and Floss encoder are trained on the \textit{ItalyPowerDemand} dataset from the UCR, as \textit{ItalyPowerDemand} exhibits daily periodicity. We use precision, recall and F1-score to measure the performance of anomaly detection. For normal settings, batch size is set to 16, Floss weight is 1,  contrastive loss weight is 0.6. For Yahoo(\textit{Cold-start})and KPI(\textit{Cold-start}), Batch size is 16, Floss weight is 1,  contrastive loss weight is 1.


\subsubsection{Experimental Results} The anomaly detection performance of TS2Vec-Floss, TS2Vec, and a strong unsupervised learning baseline SR~\cite{ren2019time} are presented in Table~\ref{tab:anomaly}. In the normal setting, Floss improves the F1 score by 1.19\% on the \textit{Yahoo} dataset and 1.08\% on the \textit{KPI} dataset compared to TS2Vec. This indicates that Floss is more sensitive to outliers in time series, as it captures periodic dynamics and expresses fine-grained information through hierarchical pooling. In the cold start setting, the improvement of Floss on both datasets is even more noticeable (about 10\% on F1 score), demonstrating its ability to capture general periodic invariance with strong transferability.

\begin{table}[!t]
    \centering
    \caption{Univariate time series anomaly detection results.}
    \begin{tabular}{clclclc}
    \toprule   
        Dataset& ~ & Yahoo & ~ & ~ & KPI & ~  \\ 
    \midrule 
       Metric & F1 & Prec. & Rec.  ~ & F1 & Prec. & Rec.   \\ 
        \midrule 
        SR & 0.563 & 0.451 & 0.747 & 0.622 & 0.647 & \textbf{0.598}  \\ 
        TS2Vec & 0.745& 0.729& 0.762 & 0.677 & 0.929 & 0.533  \\ 
        TS2Vec-FLoss&\textbf{0.754} & \textbf{0.752} & \textbf{0.763} & \textbf{0.799} & \textbf{0.946} & 0.559  \\ 
        \midrule 
        \textit{Cold-start:} & ~ \\
        SR & 0.529 & 0.404 & 0.765 & 0.666 & 0.637 &\textbf{ 0.697}  \\ 
        TS2Vec & 0.726 & 0.692 & 0.763 & 0.676 & 0.907 & 0.540  \\  
        TS2Vec-FLoss & \textbf{0.734} & \textbf{0.706} & \textbf{ 0.769} & \textbf{0.741} & \textbf{0.942} & 0.594 \\    
        
    \bottomrule 
    \end{tabular}
    \label{tab:anomaly}
\end{table}

\begin{figure}
  \centering
  \begin{subfigure}[b]{0.22\textwidth}
    \centering
    \includegraphics[width=\textwidth]{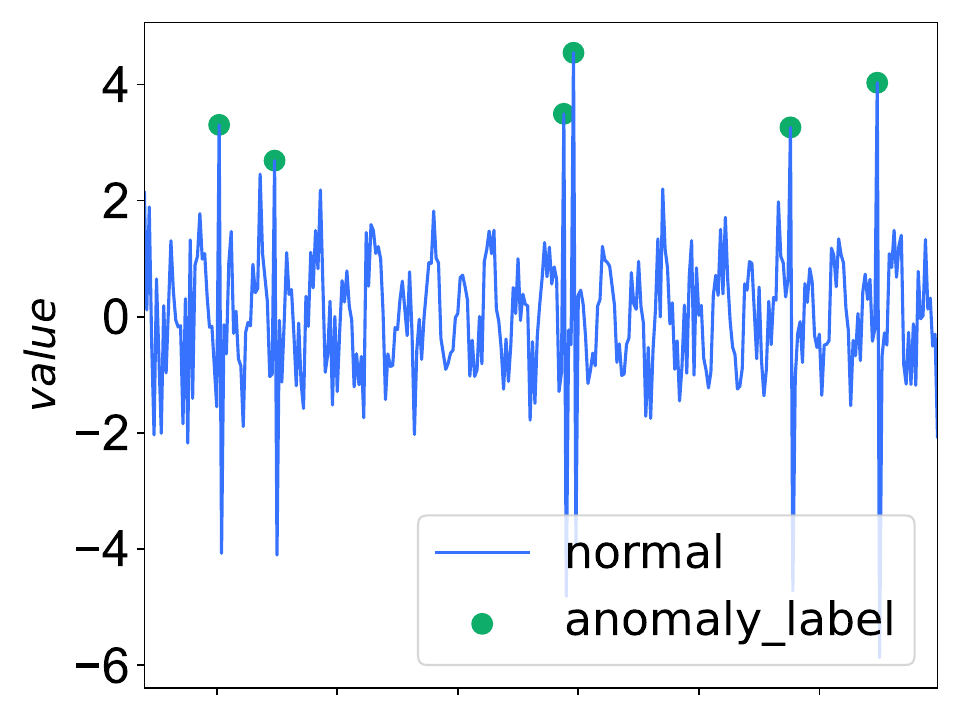}
    \caption{Anomaly detection labels of Yahoo}
    \label{fig:ano1}
  \end{subfigure}
  \hfill
   \begin{subfigure}[b]{0.22\textwidth}
    \centering
    \includegraphics[width=\textwidth]{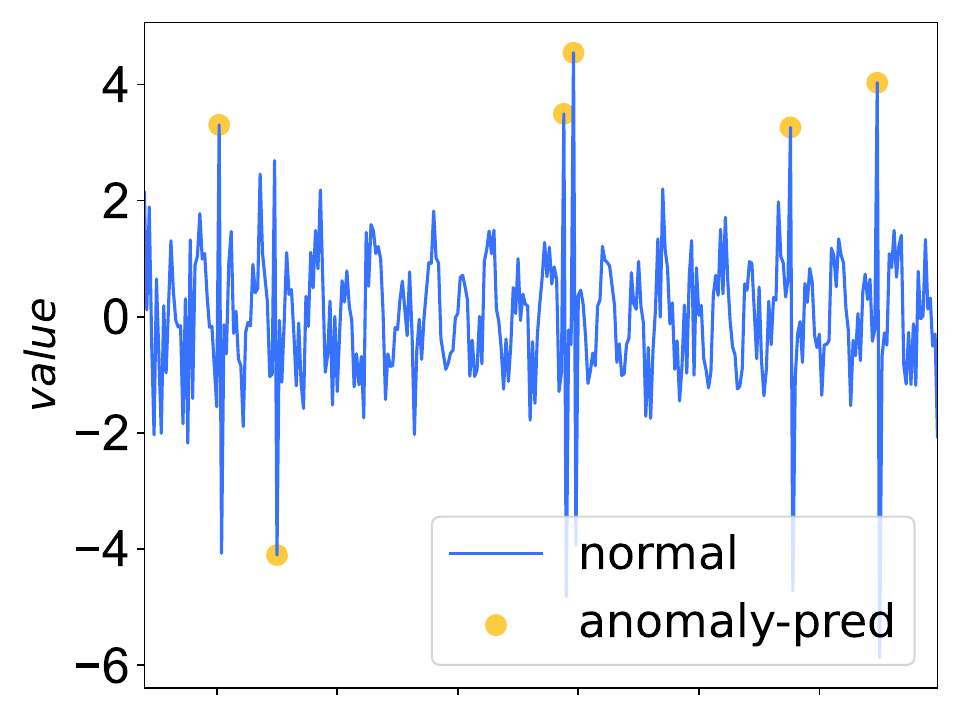}
    \caption{Anomaly predicted by Floss of Yahoo}
    \label{fig:ano2}
  \end{subfigure}
  \caption{Anomaly detection results}
  \label{fig:anomaly}
\end{figure}

We also provide visualizations of the anomaly detection performed by Floss in Figure~\ref{fig:anomaly}. In both examples, we observe that Floss accurately identifies all anomalies. It is worth noting that the only negative result obtained by Floss is still in close proximity to the corresponding ground truth anomaly time point.

\subsection{Unsupervised Anomaly Detection with other Models}

\subsubsection{Experimental Setup} In this study, we perform experiments on two extensively utilized anomaly detection datasets: MSL (Mars Science Laboratory
rover)~\cite{hundman2018detecting} and SMD (Server Machine Dataset)~\cite{su2019robust}. We selected these two datasets due to their pronounced periodic patterns. Adhering to the pre-processing techniques outlined in Anomaly Transformer~\cite{xu2021anomaly}, we divided the dataset into sequential, non-overlapping segments using a sliding window approach. Subsequently, we employed a deep learning model to reconstruct the input samples, with the resulting reconstruction error serving as the inherent anomaly indicator. To ensure equitable comparisons, we solely modified the base models for reconstruction, employing the conventional reconstruction error as the standardized anomaly criterion across all experiments. Each dataset consists of training and testing subsets, with validation subsets identical to the testing subsets. Anomalies are only labeled within the testing subset. The calculation of Floss is integrated into the `anomaly\_detection` method of each model. Initially, we perform periodicity detection on the input data and extract periodic segments. Subsequently, we extract features from these segments and calculate the Floss. Finally, the Floss is incorporated into the model training process. Throughout these experiments, the Floss weight is set to 1, and the reconstruction loss weight is set to 0.3. 

After training the model, we analyze the training data within a gradient-free context. For each data batch, we employ the trained model to reconstruct it and calculate the reconstruction error scores. To establish the anomaly threshold, we aggregate scores from both the training and test datasets. This combined score assists in determining the threshold, based on a predefined anomaly ratio. Subsequently, we compare the test data scores with the threshold to identify anomalies. Scores exceeding the threshold are classified as anomalies, while those falling below it are categorized as normal. 

\subsubsection{Models Improved by Floss.} We consider four notable multivariate time series forecasting models as featured in \cite{wu2023timesnet}: 1).FEDformer~\cite{zhou2022fedformer}: This model combines a Transformer architecture with the seasonal-trend decomposition method. 2).TimesNet~\cite{wu2023timesnet}: It employs Fast Fourier Transform (FFT) to convert the time series into a 2D representation, utilizing CNNs as the foundational framework. 3). Reformer~\cite{kitaev2019reformer}: This variant of the Transformer replaces the conventional dot-product attention mechanism with a locality-sensitive hashing approach. 4). A conventional Transformer~\cite{vaswani2017attention}.

\subsubsection{Results} Table~\ref{tab:anomaly2} illustrates that Floss continues to enhance anomaly detection performance, yielding improvements for the selected model in most instances. We have summarized some intriguing observations as follows: 1). Floss demonstrates a notable improvement in the F1 scores for nearly all models, with the exception being TimesNet on the SMD dataset. This discrepancy could potentially arise from TimesNet's adept utilization of periodic information through its inherent FFT block. 2). Remarkably, the Transformer-Floss combination attains the highest F1 score on the MSL dataset, surpassing even the more intricate TimesNet model. This outcome suggests that Floss can imbue a simpler model with robust time series processing capabilities, offering valuable insights for designing models in the context of anomaly detection tasks.

\begin{table*}[!t]
    \centering
    \caption{Anomaly detection task. We calculate the accuracy, precision, recall and F1 scores for each dataset.}
        \label{tab:anomaly2}
    \begin{tabular}{c|cccc|cccc}
    \toprule   
        Dataset&\multicolumn{4}{c}{MSL}& \multicolumn{4}{c}{SMD} \\ 
    \midrule 
       Metric & Acc. & Prec. & Rec. & F1 & Acc. & Prec. & Rec. & F1    \\ 
        \midrule 
        FEDformer & 0.9673 & 0.7714 & 0.7679 & 0.7857 & 0.9763 & 0.7732 & 0.6094 & 0.6816  \\ 
        FEDformer-Floss & 0.9651 &\textbf{0.9059} & 0.7465 & \textbf{0.8185} & \textbf{0.9781} & \textbf{0.7846} & \textbf{0.6508}  & \textbf{0.7114}\\  
        \midrule 
        TimesNet & 0.9647 & 0.8955 & 0.7529 & 0.8180 & 0.9877 & 0.8788 & 0.8154 & 0.8459  \\ 
        TimesNet-Floss & \textbf{0.9648} &\textbf{0.8959} & \textbf{0.7541} & \textbf{0.8187} & 0.9867 & 0.8684 & 0.8008  & 0.8332\\  
        \midrule 
        Reformer & 0.9638 & 0.9014 & 0.7372 & 0.8111 & 0.9780 & 0.7832 & 0.6524 & 0.7118  \\ 
        Reformer-Floss & \textbf{0.9647} &\textbf{0.9055} & \textbf{0.7430} & \textbf{0.8163} & \textbf{0.9781} & 0.7832 & \textbf{0.6538}  & \textbf{0.7127}\\  
        \midrule 
         Transformer & 0.9634 & 0.8977 & 0.7366 & 0.8093 & 0.9780 & 0.7832 & 0.6524 & 0.7118  \\ 
        Transformer-Floss & \textbf{0.9652} &\textbf{0.9062} & \textbf{0.7470} & \textbf{0.8189} & \textbf{0.9781} & 0.7828 & \textbf{0.6537}  & \textbf{0.7125}\\          
    \bottomrule 
    \end{tabular}
    \end{table*}

\begin{figure}
  \centering
  \includegraphics[width=0.9\linewidth]{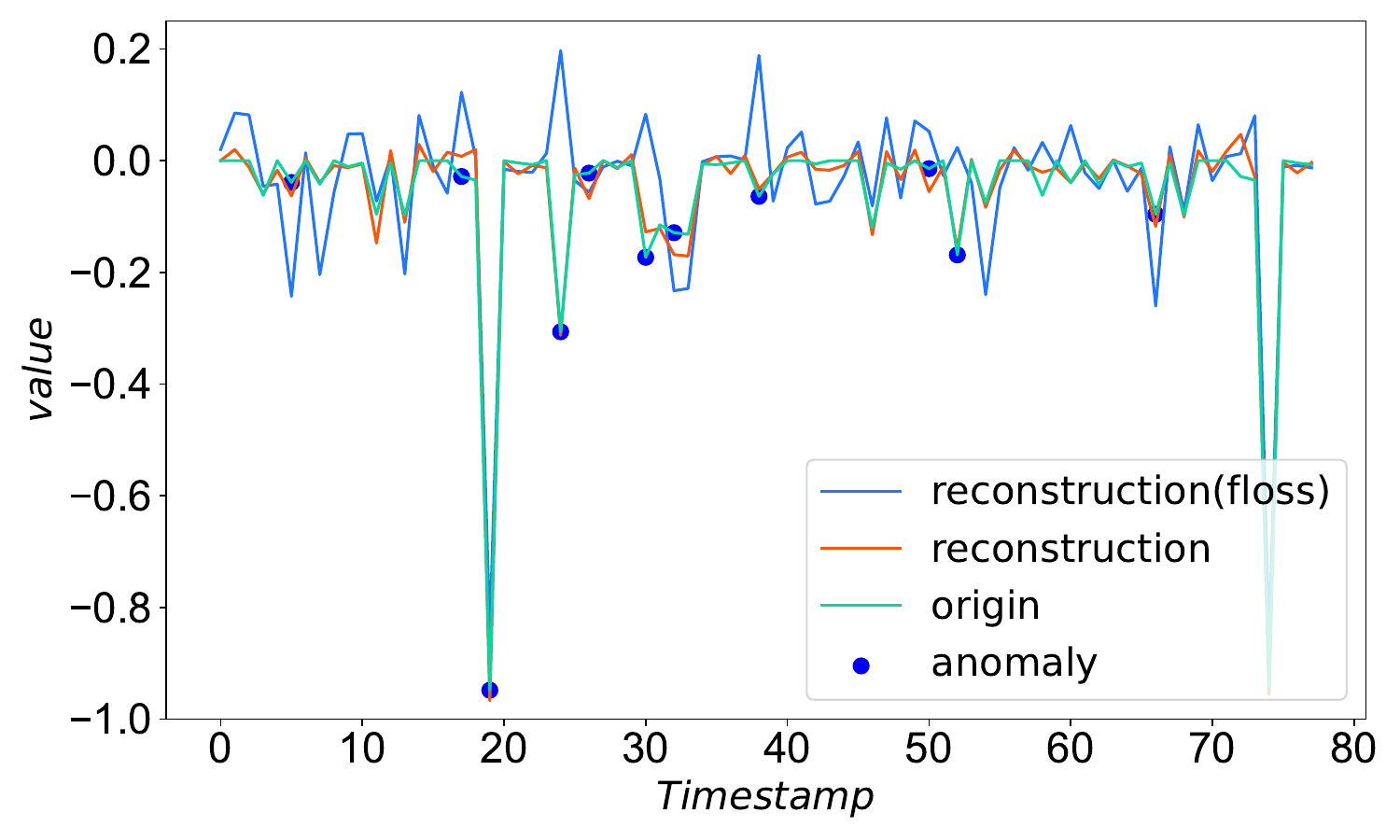}
  \caption{Visualization of the reconstruction, true value and anomaly.}
  \label{fig:anomaly2}
\end{figure}

We demonstrate the reconstruction effects after incorporating Floss into the Transformer in Figure~\ref{fig:anomaly2}. It can be observed that when anomalies occur, the Transformer model with Floss exhibits larger reconstruction errors compared to the regular Transformer model. Floss, to some extent, preserves the consistency of periodic observations in the spectrum. Many anomalies often manifest as significant changes in certain parts of the spectrum. Therefore, Floss, by maintaining the consistency of periodic spectral patterns, is advantageous for anomaly detection.

\subsection{Detailed Study of Floss}

As Floss is designed as a plug-in loss function, there can be various instances with different implementation choices for each module. In this section, a comprehensive analysis and comparison of different instances of Floss are conducted. In the following discussions, we consider the simplest TS2Vec as the baseline and compare it with other variants on multivariate time series forecasting for Weather, Exchange, ILI and Ettm1. Furthermore, we employ a fixed set of hyperparameters to ensure a fair comparison. It is worth noting that some results may appear worse than those reported in Table~\ref{tab:multi_pre} because we only presented the best results in Table~\ref{tab:multi_pre}

\begin{figure}
  \centering
  \begin{subfigure}[b]{0.23\textwidth}
    \centering
    \includegraphics[width=\textwidth]{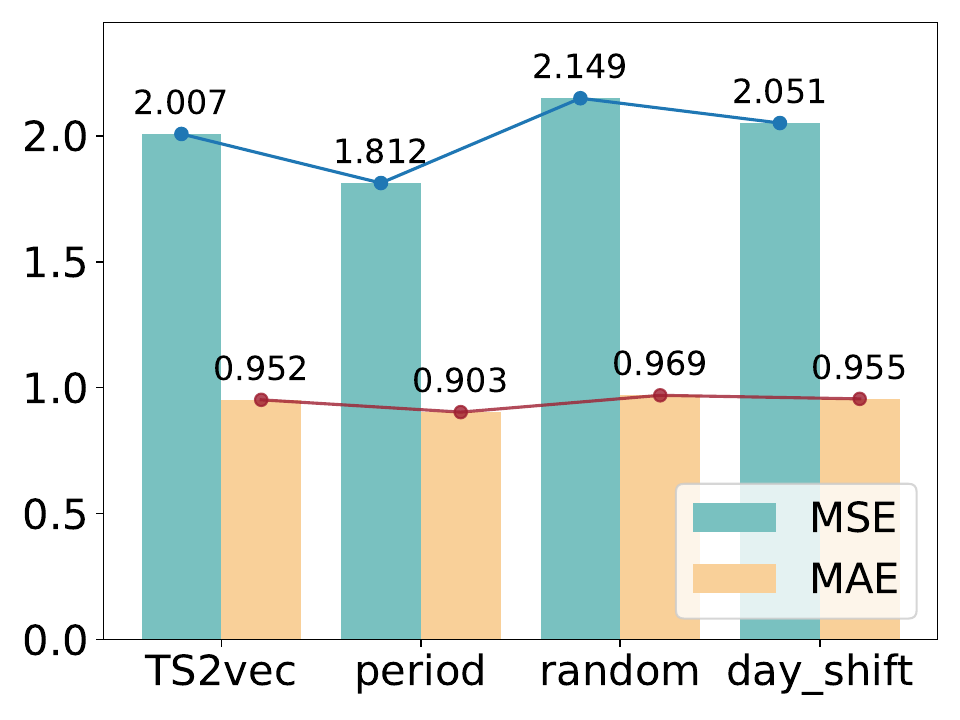}
    \caption{Effect of the period detection module on forecasting errors \\.}
    \label{fig:a1}
  \end{subfigure}
  \hfill
  \begin{subfigure}[b]{0.23\textwidth}
    \centering
    \includegraphics[width=\textwidth]{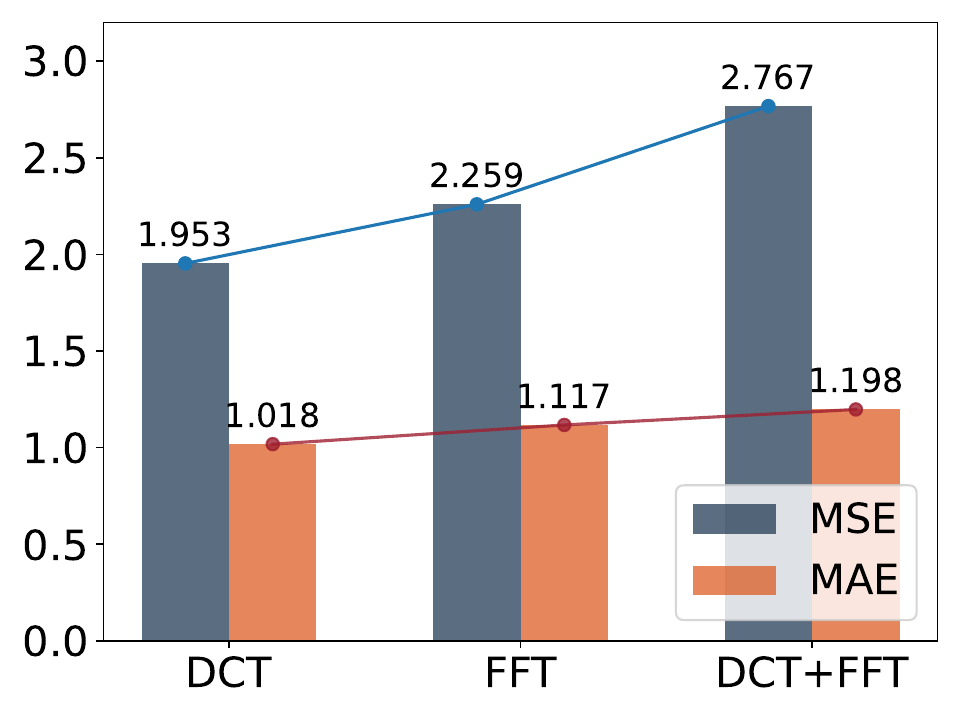}
    \caption{FFT and DCT: peridoicity is detected by FFT, DCT+FFT: periodicity is dtected by DCT. }
    \label{fig:a2}
  \end{subfigure}
  \begin{subfigure}[b]{0.23\textwidth}
    \centering
    \includegraphics[width=\textwidth]{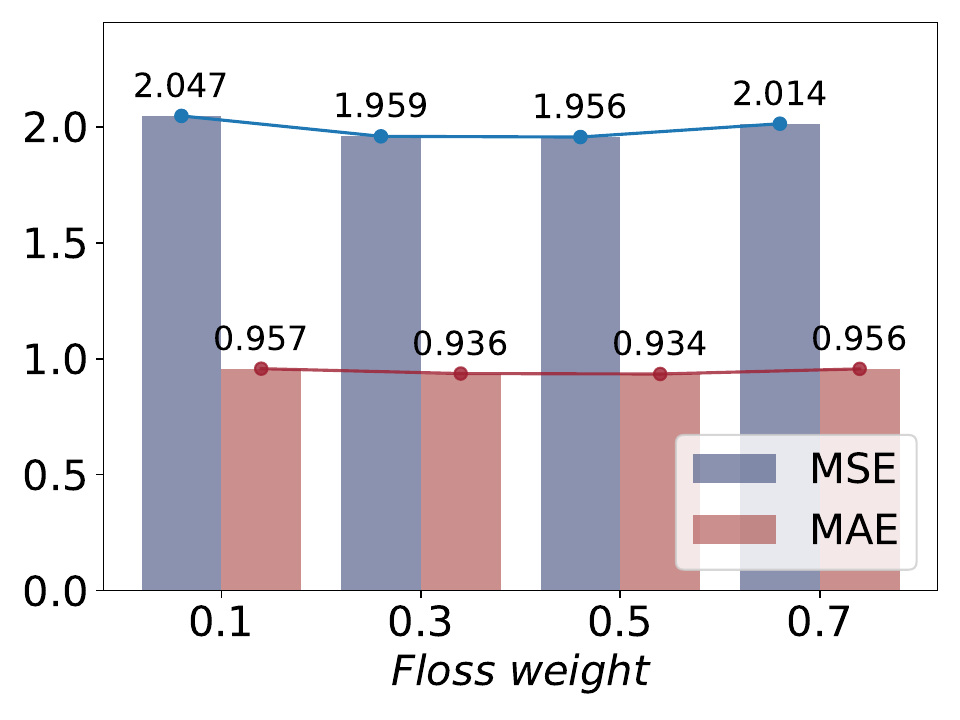}
    \caption{Effect of Floss weights on forecasting errors. }
    \label{fig:a3}
  \end{subfigure}
  \caption{Ablation results.}
  \label{fig:ablation}
\end{figure}

\subsubsection{Effects of periodic detection module}

We initiated our investigation by examining the influence of the period detection module on the model. A comparative analysis was conducted between Floss and two alternative models, namely random and day shift. In Figure~\ref{fig:a1}, 'random' signifies the utilization of random augmentation during each comparison with Floss, while 'day shift' denotes the shifting of time series by one day at each step, operating under the implicit assumption that all time series exhibit a daily periodicity. The outcomes unveiled that both models incorporating random shifting and day shifting exhibited inferior performance compared to the TS2Vec model. 

Since Floss assumes that the representation of periodic shifts is similar in the frequency domain, augmenting time series with random shifts or assuming a daily shift might not effectively capture the underlying patterns and periodic behavior. These findings suggest that considering generic shifts or assuming a specific daily pattern might overlook the nuanced dynamics of the time series. Notably, it was observed that only the model incorporating the period detection module for augmentation outperformed TS2Vec. This highlights the critical role played by the period detection module in enhancing the model's performance. 

\subsubsection{Combination of DCT and FFT}

We also conducted an examination of the combination of Fast Fourier Transform (FFT) and Discrete Cosine Transform (DCT) in relation to period identification and Floss computation. In Figure~\ref{fig:a2}, 'FFT' and 'DCT' signify the utilization of FFT and DCT for Floss calculation, respectively. Notably, both approaches employ FFT for periodic detection. On the other hand, 'DCT+FFT' indicates the application of DCT for period identification and FFT for Floss computation. Our investigation yielded noteworthy results, unveiling the significance of the combination. It was observed that employing FFT for period identification, while leveraging DCT for spectral density computation, yielded the most optimal outcomes in terms of performance. 

\subsubsection{Effects of Floss weights}

Floss operates in conjunction with the loss of other models. Our encoder is trained using a weighted sum of Floss and other loss functions. Assigning a higher weight to Floss indicates a greater reliance on capturing periodic invariances. To investigate the impact of the Floss weight, we set the contrastive loss weight of TS2Vec to 0.5 and evaluate the model's performance with different loss weights on three datasets. The results are presented in Figure~\ref{fig:a3}. The findings demonstrate the robustness of our proposed method to the choice of weight. The performance of the model remains consistent across various weight settings. However, upon closer analysis, we identify that the weight range between 0.3 and 0.5 yields the best performance. 

\subsubsection{Effects of hierarchical Floss computation}

\begin{figure}
  \centering
  \begin{subfigure}[b]{0.23\textwidth}
    \centering
    \includegraphics[width=\textwidth]{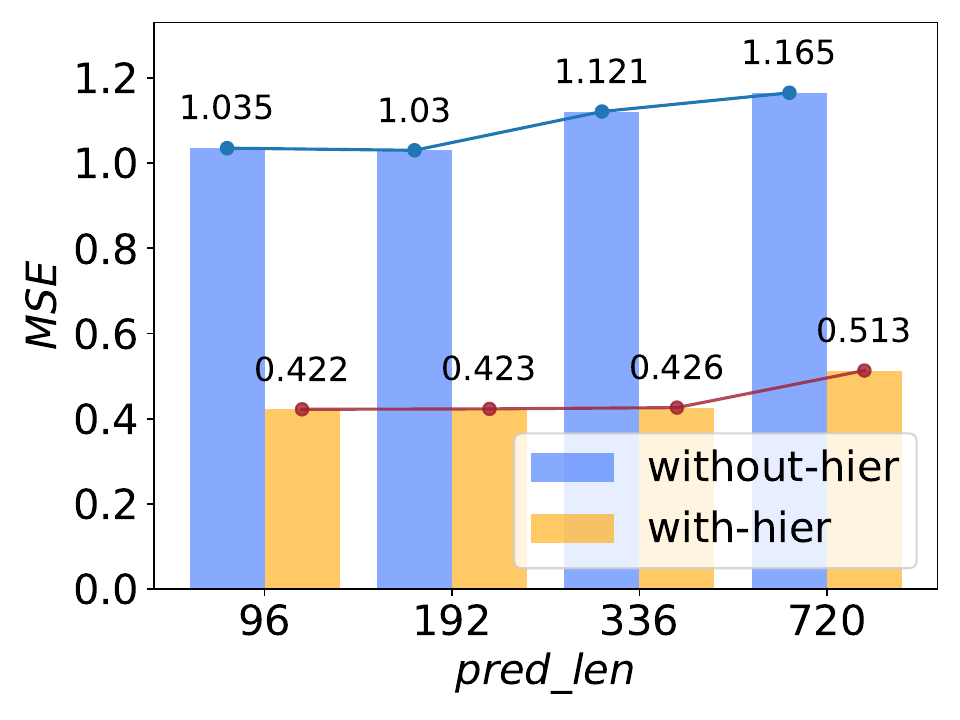}
    \caption{Effect of the hierarchical Floss computation on Electricity dataset.}
    \label{fig:h1}
  \end{subfigure}
  \hfill
  \begin{subfigure}[b]{0.23\textwidth}
    \centering
    \includegraphics[width=\textwidth]{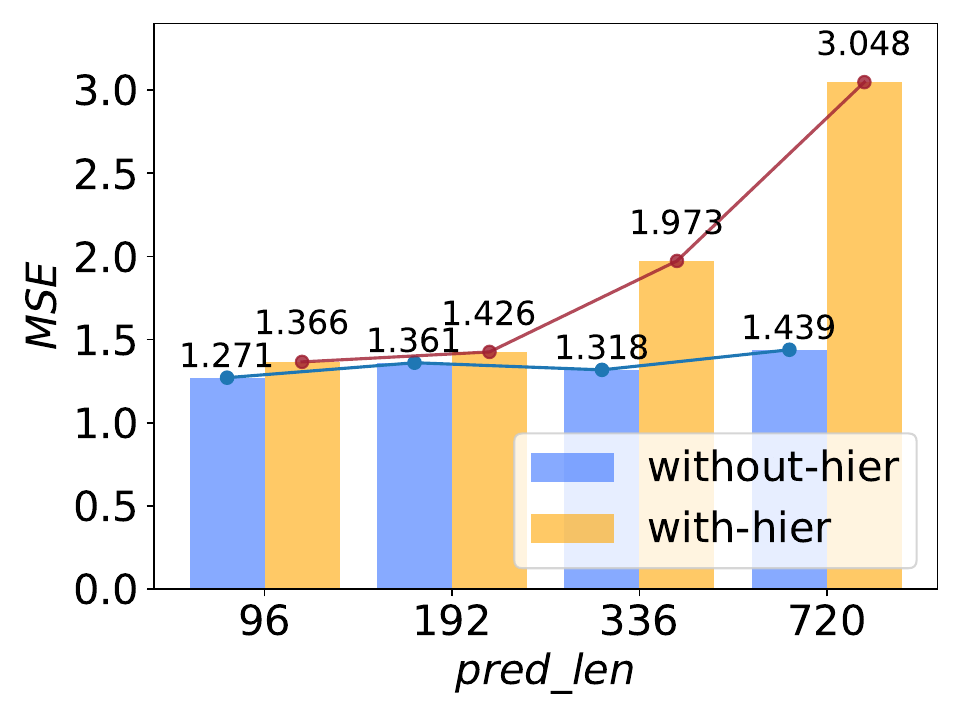}
    \caption{Effect of the hierarchical Floss computation on Weather dataset \\. }
    \label{fig:h2}
  \end{subfigure}
  \caption{Effect of hierarchical Floss computation.}
  \label{fig:hierarchical}
\end{figure}

As described in Section~\ref{hfl}, we employ a hierarchical Floss computation strategy to allocate greater weights to the low-frequency components. However, it is noteworthy that employing hierarchical Floss computation may not be necessary for all datasets. The performance comparison without hierarchical computation is presented in Figure~\ref{fig:hierarchical}. Specifically, our experimentation on the electricity dataset demonstrates a substantial enhancement in model performance when utilizing hierarchical Floss computation. In contrast, for the weather dataset, we observed that refraining from hierarchical Floss computation actually yielded superior outcomes. When employing hierarchical computation,  we tend to focus more on capturing the similarities in the low-frequency components. On the other hand, without employing hierarchical computation, we treat all frequency components equally, including the high-frequency components. This observation suggests that in datasets such as weather, after undergoing periodic variations, the abstract representation of the high-frequency components remains relatively unchanged. Preserving all frequencies becomes more effective for such data. 

Moreover, we observed a significant improvement in long-term forecasting performance when hierarchical Floss computation was not employed for weather dataset. This finding suggests that for the weather dataset, the long-term variation trend may be concealed within the unchanged high-frequency components under periodic shifts. These phenomena call for further in-depth research to design more robust models capable of capturing these patterns.

\subsubsection{Representation visualization} 
\begin{figure}[!htb]
\centering
  \centering
  \begin{subfigure}[b]{0.235\textwidth}
    \centering
    \includegraphics[width=\textwidth]{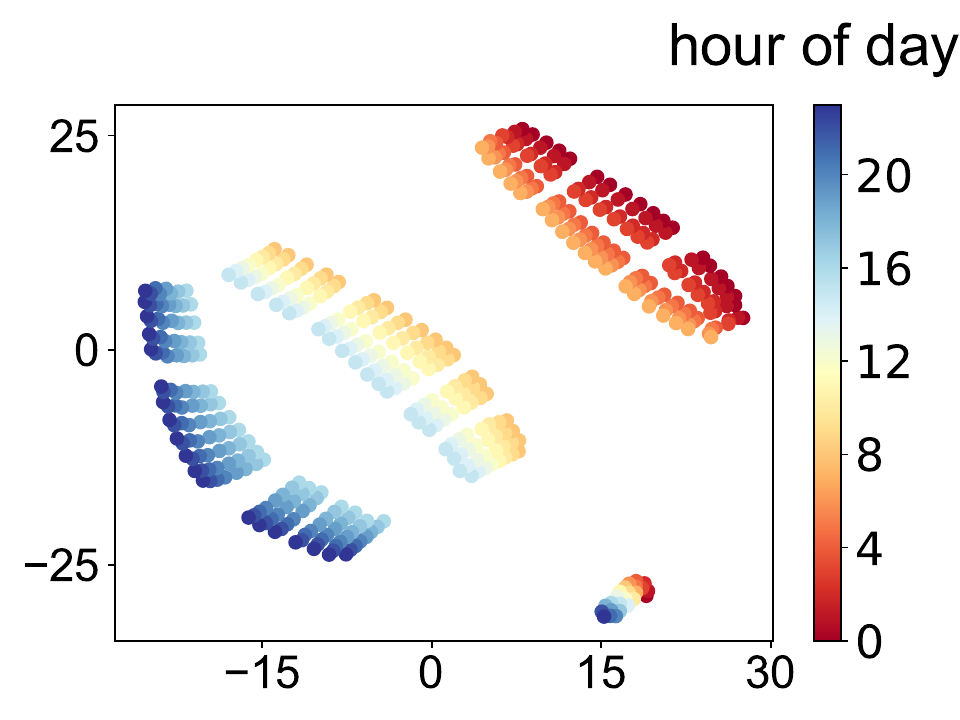}
    \caption{Floss representations on \textit{Electricity}}
    \label{fig:tsne1}
  \end{subfigure}
  \hfill
   \begin{subfigure}[b]{0.235\textwidth}
    \centering
    \includegraphics[width=\textwidth]{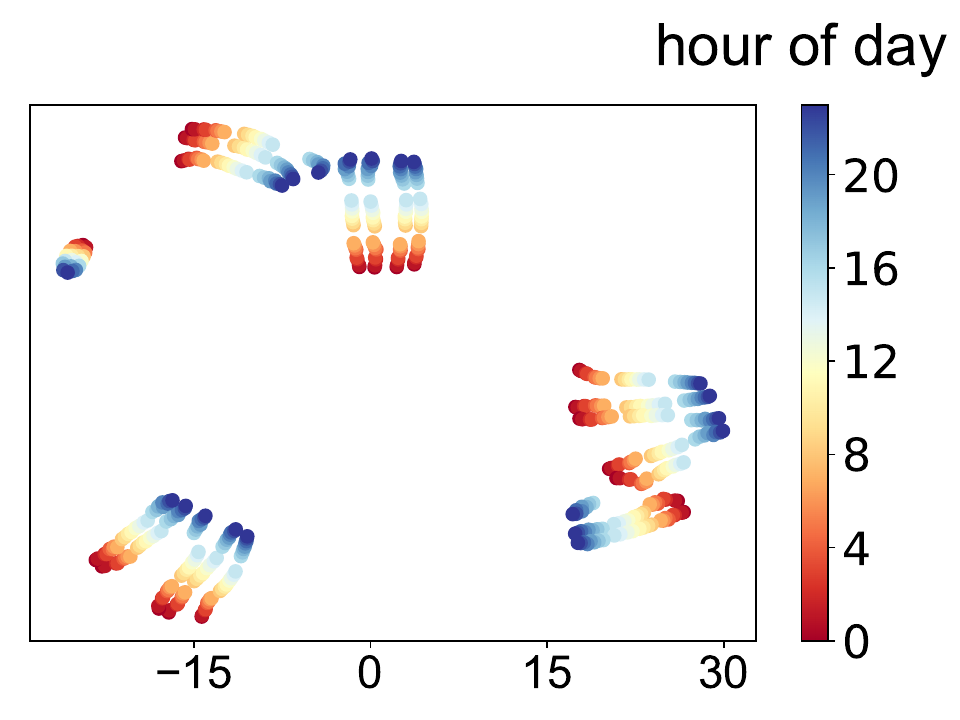}
    \caption{TS2Vec representations on \textit{Electricity}}
    \label{fig:tsne2}
  \end{subfigure}
    \begin{subfigure}[b]{0.235\textwidth}
    \centering
    \includegraphics[width=\textwidth]{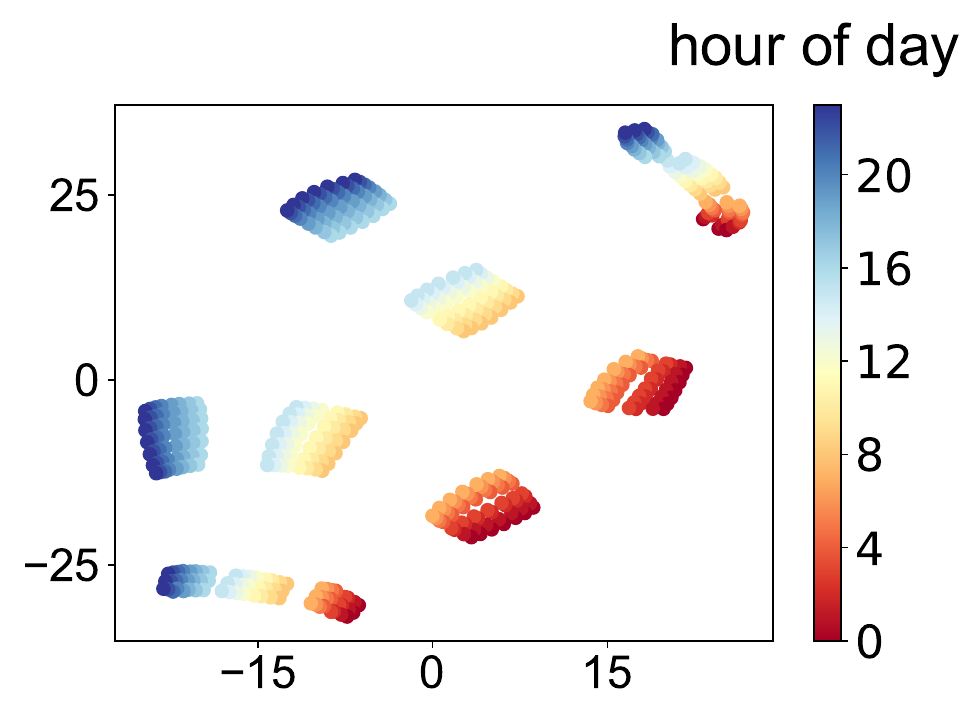}
    \caption{Floss representations on \textit{ETTh1}}
    \label{fig:tsne3}
  \end{subfigure}
  \hfill
   \begin{subfigure}[b]{0.235\textwidth}
    \centering
    \includegraphics[width=\textwidth]{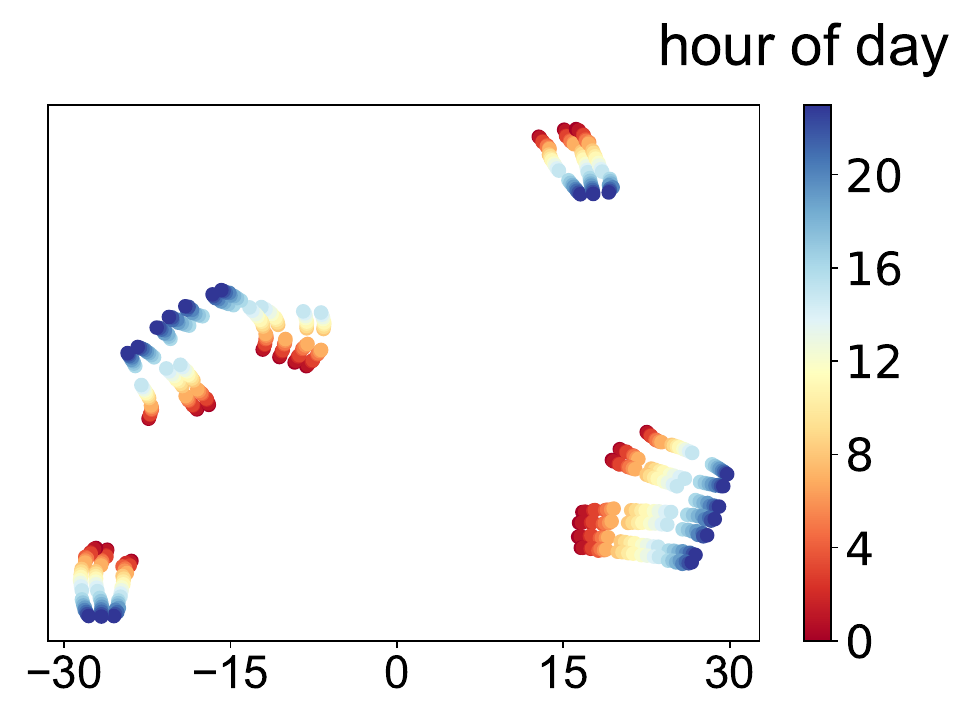}
    \caption{TS2Vec representations on \textit{ETTh1}}
    \label{fig:tsne4}
  \end{subfigure}
\caption{T-SNE visualizations of the learned representations of TS2Vec-Floss and TS2Vec on \textit{Electricity} and \textit{ETTh1}. Different colors represent different hours of day.}
\label{fig:tsne}
\end{figure}

Figure~\ref{fig:tsne} displays the t-SNE embedding of TS2Vec-Floss and Floss on nine consecutive days of the \textit{Electricity} and \textit{ETTh1} datasets. These datasets are known to exhibit pronounced daily periodicity. Consequently, the automatic periodic detection module is anticipated to capture this strong periodic pattern. In this visualization, the model with Floss produces a more periodic cloud structure, characterized by a reduced presence of easily distinguishable hour-of-day groupings.

\subsubsection{Accuracy of Periodicity Detection}

\begin{figure}
  \centering
\includegraphics[width=0.5\linewidth]{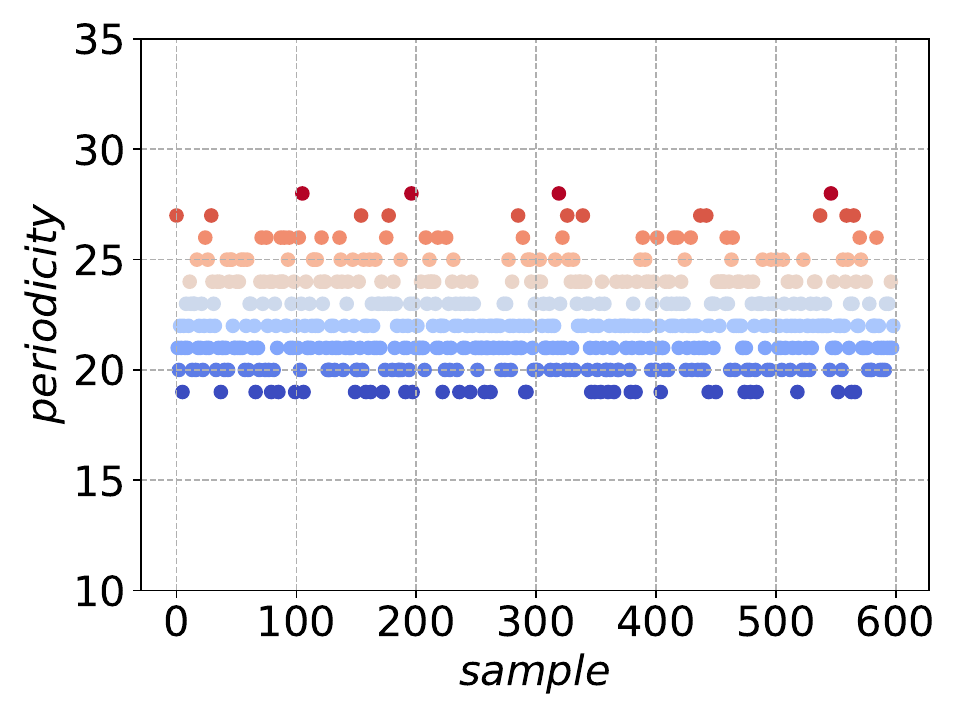}
  \caption{Periodicity Detection Results with informer-Floss for Electricity.}
  \label{fig:period}
\end{figure}

We provide a case study (informer-Floss for Electricity) of the periodicity detection module in Figure~\ref{fig:period}. We can observe that Floss can accurately capture the periodicities. Moreover, most of the detected periodicities are approximately equal to 24 hours (1 day), which supports our motivation in adopting automatic periodicity detection for representation learning.

\section{Conclusion}
\label{sec:con}

In this study, we addressed the challenge of effectively capturing periodic or quasi-periodic dynamics present in real-world time series data using deep learning approaches. While deep learning has shown impressive performance in various application domains, it often struggles to adequately represent the underlying periodic behaviors in time series data. To bridge this gap, we introduced an unsupervised method called Floss. Floss is designed to automatically detect major periodicities in time series data and utilizes periodic shift and spectral density similarity measures to learn meaningful representations with periodic consistency in the frequency domain. By seamlessly incorporating Floss into supervised, semi-supervised, and unsupervised learning frameworks, we demonstrated its versatility and ability to enhance time series analysis tasks.

Our extensive experiments on common time series analysis tasks showcased the effectiveness of Floss. It outperformed state-of-the-art deep learning models, validating its capability to automatically discover periodic dynamics. The results underscore the importance of considering domain-specific knowledge about periodic behaviors to enrich the learned representations in deep learning models. 

For future work, exploring advanced modeling techniques that can effectively capture the hidden long-term patterns in complex data such as weather data remains a promising direction. In the weather dataset, we observed a significant improvement in long-term forecasting performance when hierarchical Floss computation was not employed. his finding suggests that for some datasets, the long-term variation trend may be concealed within the unchanged high-frequency components under periodic shifts. For future work, exploring advanced modeling techniques that can effectively capture the hidden long-term patterns in weather data remains a promising direction. This may involve the incorporation of domain-specific knowledge, such as external factors, to enhance the modeling process. Furthermore, extending the research to consider more complex and dynamic scenarios, such as time series prediction under extreme weather events, could present new challenges and opportunities for advancing time series analysis. Floss solely addresses the frequency domain similarity of the model concerning temporal periodicity. Integrating state-of-the-art techniques, such as Graph Neural Networks (GNNs) and graph spectral analysis, holds promise for modeling inter time series invariance and optimizing time series analysis performance. By leveraging GNNs and graph spectral analysis, we can gain a deeper understanding of the relationships between multiple time series, capturing intricate temporal dependencies and interdependencies among time series.

\begin{acks}
 This work was supported by National Natural Science Foundation of Sichuan Province (Grant No.2023NSFSC1423), the Tianfu Emei Plan of Sichuan Province, and the Fundamental Research Funds for the Central Universities.
\end{acks}


\bibliographystyle{ACM-Reference-Format}
\bibliography{sample}

\end{document}